\documentclass[sigconf]{acmart}
\usepackage{times}
\usepackage{latexsym}
\usepackage{xcolor}
\usepackage{float}
\usepackage{tabularx,booktabs} 
\usepackage{multirow}
\usepackage{xspace}

\usepackage{amsmath}

\usepackage{enumitem}

\usepackage{graphicx}
\usepackage{subcaption}

\usepackage{microtype}

\newenvironment{remark}[1][Remark]{\begin{trivlist}
\item[\hskip \labelsep {\bfseries #1}]}{\end{trivlist}}

\newcommand{\nop}[1]{}

\usepackage{amsmath}

\usepackage[disable]{todonotes} % uncomment this line and comment out the above line to disable notes
\makeatletter
\newcommand*\iftodonotes{\if@todonotes@disabled\expandafter\@secondoftwo\else\expandafter\@firstoftwo\fi}  % defines \iftodonotes{<true>}{<false>}, thanks to https://tex.stackexchange.com/questions/126559/conditional-based-on-packageoption
\makeatother

\newlength{\extramargin}
\usepackage{calc}
\usepackage[hang]{footmisc}
\usepackage{lipsum}

\iftodonotes{\usepackage[paperwidth=\paperwidth+\extramargin*2,marginparwidth=\marginparwidth+\extramargin,width=\textwidth,height=\textheight]{geometry}}{}
\newcommand{\E}{$\mathcal{E\,}$}
\renewcommand{\L}{$\mathcal{L\,}$}
\newcommand{\R}{$\mathcal{R\,}$}
\newcommand{\Cls}{$\mathcal{C\,}$}

\newcommand{\Problem}{KBQA\xspace}
\newcommand{\OurDataset}{\textsc{GrailQA}\xspace}
\newcommand{\GoogleKG}{\textsc{Google Knowledge Graph}\xspace}
\newcommand{\Freebase}{\textsc{Freebase}\xspace}
\newcommand{\DBpedia}{\textsc{DBpedia}\xspace}

\newcommand{\Commons}{\textsc{Commons}\xspace}
\newcommand{\WebQ}{\textsc{WebQ}\xspace}
\newcommand{\WebQSP}{\textsc{WebQ}\xspace}
\newcommand{\ComplexQ}{\textsc{ComplexWebQ}\xspace}
\newcommand{\GraphQ}{\textsc{GraphQ}\xspace}
\newcommand{\SimpleQ}{\textsc{SimpleQ}\xspace}
\newcommand{\QALD}{\textsc{QALD}\xspace}
\newcommand{\QuAD}{\textsc{LC-QuAD}\xspace}
\newcommand{\CFQ}{\textsc{CFQ}\xspace}
\newcommand{\Overnight}{\textsc{Overnight}\xspace}
\newcommand{\Ranking}{\textsc{Ranking}\xspace}
\newcommand{\Transduction}{\textsc{Transduction}\xspace}
\newcommand{\QGG}{\textsc{QGG}\xspace}

\newenvironment{entityfont}{\fontfamily{phv}\selectfont\footnotesize}{\par}
\newenvironment{relationfont}{\fontfamily{phv}\selectfont\footnotesize}{\par}
\DeclareTextFontCommand{\textentity}{\entityfont}
\DeclareTextFontCommand{\textrelation}{\relationfont}
\newcommand\nl[1]{{\it``#1''}} % Natural language
\newcommand\wl[1]{{\fontfamily{phv}\selectfont\footnotesize#1}}

\newcommand{\iid}{i.i.d.\xspace}
\newcommand{\IID}{I.I.D.\xspace}

\setcopyright{iw3c2w3}
\begin{document}

\copyrightyear{2021}
\acmYear{2021}
\acmConference[WWW '21]{Proceedings of the Web Conference 2021}{April 19--23,
2021}{Ljubljana, Slovenia}
\acmBooktitle{Proceedings of the Web Conference 2021 (WWW '21), April 19--23, 2021,
Ljubljana, Slovenia}
\acmPrice{}
\acmDOI{10.1145/3442381.3449992}
\acmISBN{978-1-4503-8312-7/21/04}

%%
%% The "title" command has an optional parameter,
%% allowing the author to define a "short title" to be used in page headers.
\title{Beyond \IID: Three Levels of Generalization for Question Answering on Knowledge Bases}

%%
%% The "author" command and its associated commands are used to define
%% the authors and their affiliations.
%% Of note is the shared affiliation of the first two authors, and the
%% "authornote" and "authornotemark" commands
%% used to denote shared contribution to the research.
% \author{Ben Trovato}
% \authornote{Both authors contributed equally to this research.}
% \email{trovato@corporation.com}
% \orcid{1234-5678-9012}
% \author{G.K.M. Tobin}
% \authornotemark[1]
% \email{webmaster@marysville-ohio.com}
% \affiliation{%
%   \institution{Institute for Clarity in Documentation}
%   \streetaddress{P.O. Box 1212}
%   \city{Dublin}
%   \state{Ohio}
%   \postcode{43017-6221}
% }
\author{Yu Gu}
\affiliation{The Ohio State University}
\email{gu.826@osu.edu}
\author{Sue Kase}
\affiliation{U.S. Army Research Laboratory}
\email{sue.e.kase.civ@mail.mil}
\author{Michelle T. Vanni}
\affiliation{U.S. Army Research Laboratory}
\email{michelle.t.vanni.civ@mail.mil}
\author{Brian M. Sadler}
\affiliation{U.S. Army Research Laboratory}
\email{brian.m.sadler6.civ@mail.mil}
\author{Percy Liang}
\affiliation{Stanford University}
\email{pliang@cs.stanford.edu}
\author{Xifeng Yan}
\affiliation{University of California, Santa Barbara}
\email{xyan@cs.ucsb.edu}
\author{Yu Su}
\affiliation{The Ohio State University}
\email{su.809@osu.edu}

%%
%% By default, the full list of authors will be used in the page
%% headers. Often, this list is too long, and will overlap
%% other information printed in the page headers. This command allows
%% the author to define a more concise list
%% of authors' names for this purpose.
% \renewcommand{\shortauthors}{Anonymous}

\begin{abstract}
Existing studies on question answering on knowledge bases (\Problem) mainly operate with the standard \iid assumption, i.e., training distribution over questions is the same as the test distribution. However, \iid may be neither reasonably achievable nor desirable on large-scale KBs because 1) true user distribution is hard to capture and 2) randomly sampling training examples from the enormous space would be highly data-inefficient. Instead, we suggest that \Problem models should have three levels of built-in generalization: \emph{\iid}, \emph{compositional}, and \emph{zero-shot}. To facilitate the development of \Problem models with stronger generalization, we construct and release a new large-scale, high-quality dataset with 64,331 questions, \OurDataset, and provide evaluation settings for all three levels of generalization. 
In addition, we propose a novel BERT-based \Problem model. The combination of our dataset and model enables us to thoroughly examine and demonstrate, for the first time, the key role of pre-trained contextual embeddings like BERT in the generalization of \Problem.\footnote{Data and leaderboard: http://dki-lab.github.io/GrailQA/\\ Code: https://github.com/dki-lab/GrailQA}

% It presents an array of unique challenges due to its broad coverage and large search space. However, despite substantial development in the early stage, benchmark results on OpenSP seem to have soon plateaued and further improvement has turned out to be difficult. In this paper, we carefully examine the OpenSP problem and provide (1) a large-scale, high-quality dataset consisting of 64,616 compositional questions that captures many of the challenges of OpenSP, (2) new baseline models with strong cross-domain generalizability using pre-trained contextual embeddings such as BERT, and (3) in-depth discussion and insights that suggest venues of future improvement on OpenSP. We also show that our dataset serves as a valuable pre-training corpus for OpenSP in general, and models pre-trained on our dataset can generalize to other datasets like \WebQ in both few-shot and zero-shot fashion. 
\end{abstract}

%%
%% The code below is generated by the tool at http://dl.acm.org/ccs.cfm.
%% Please copy and paste the code instead of the example below.
%%
% \begin{CCSXML}
% <ccs2012>
%   <concept>
%       <concept_id>10010147.10010178.10010179.10010186</concept_id>
%       <concept_desc>Computing methodologies~Language resources</concept_desc>
%       <concept_significance>500</concept_significance>
%       </concept>
%  </ccs2012>
% \end{CCSXML}

% \ccsdesc[500]{Computing methodologies~Language resources}

% \keywords{knowledge base, question answering, semantic parsing, datasets}

%% A "teaser" image appears between the author and affiliation
%% information and the body of the document, and typically spans the
%% page.
\begin{teaserfigure}
  \includegraphics[width=\textwidth]{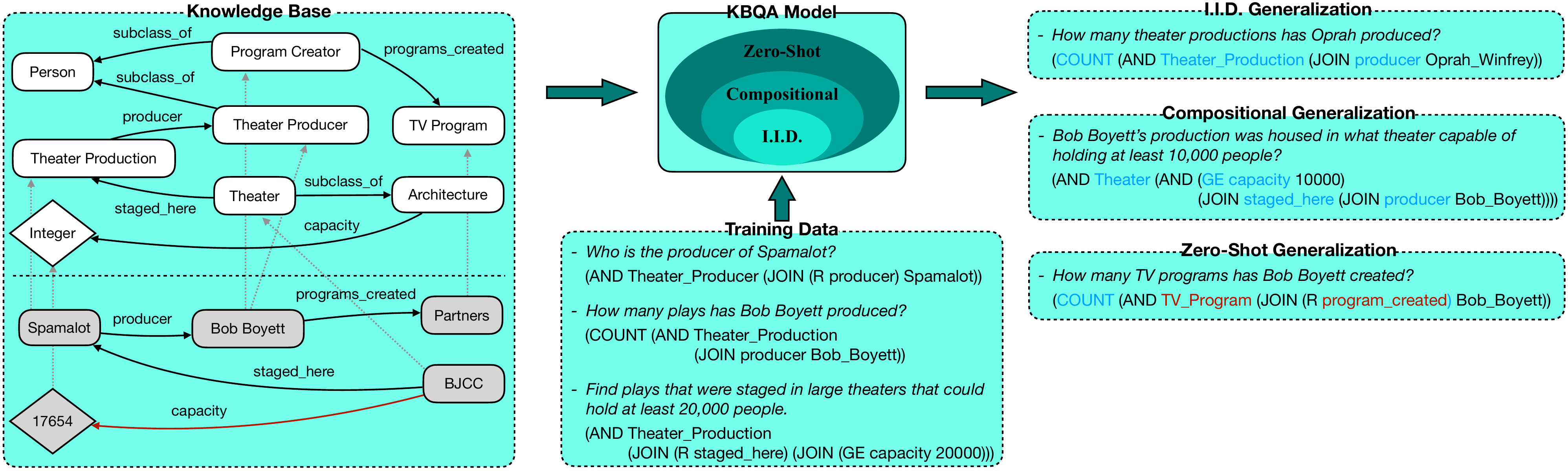}
  \caption{On large-scale KBs, collecting sufficient training data for \Problem to ensure \iid distribution at test time is very difficult, if possible at all. We argue that practical \Problem models should have three levels of \emph{built-in generalization} rather than solely relying on training data: (1) \emph{\iid generalization} to questions following the training distribution, (2) \emph{compositional generalization} to novel compositions of schema items seen in training (marked blue), and (3) \emph{zero-shot generalization} to unseen schema items or even domains (marked red). Our definition of generalization is based on the underlying logical forms (shown as S-expressions). Orthogonally, as illustrated by the examples, \Problem models should also have strong generalization to linguistic variation. Figure best viewed in color.}
%   \Description{Enjoying the baseball game from the third-base
%   seats. Ichiro Suzuki preparing to bat.}
  \label{fig:intro}
\end{teaserfigure}

%%
%% This command processes the author and affiliation and title
%% information and builds the first part of the formatted document.
\maketitle

\section{Introduction}

% The Internet and search engines have greatly improved information accessibility\nop{, which promotes the image that the world is at our fingertips}.\nop{ The Linked Open Data initiative aims to build a semantic layer on top of the web, while billions of intelligent devices are being connected in the Internet of Things.} However, as the available data and services rapidly become more massive and heterogeneous, standing in stark contrast to the spreading demand is the growing gap between end users and the computing world: Accessing each kind of data or service induces a learning curve for mastering the GUI or sometimes even a programming language, which in turn results in siloed and limited access~\cite{techcrunch2015smartphone}.

Question answering on knowledge bases (\Problem) has emerged as a promising technique to provide unified, user-friendly access to knowledge bases (KBs) and shield users from the heterogeneity underneath~\cite{berant-etal-2013-semantic,kwiatkowski-etal-2013-scaling,cai-yates-2013-large,qald9}. As the scale and coverage of KBs increase, \Problem is becoming even more important due to the increasing difficulty of writing structured queries like SPARQL.\footnote{\Freebase contains over 100 domains, 45 million entities, and 3 billion facts. \GoogleKG has amassed over 500 billion facts about 5 billion entities~\cite{googlekg2020}.}

There has been an array of datasets and models for \Problem in recent years~\cite{berant-etal-2013-semantic,su-etal-2016-generating,talmor-berant-2018-web,qald9,trivedi2017lc,yih-etal-2015-semantic,dong-lapata-2016-language,Abujabal,bhutani2019learning}. Most existing studies are (implicitly) focused on the \emph{\iid}\ setting, that is, assuming training distribution is representative of the true user distribution and questions at test time will be drawn from the same distribution. While it is standard for machine learning, it could be problematic for \Problem on large-scale KBs. First, it is very difficult to collect sufficient training data to cover all the questions users may ask due to the broad coverage and combinatorial explosion. Second, even if one strives to achieve that, e.g., by iteratively annotating all user questions to a deployed \Problem system, it may hurt user experience because the system would keep failing on new out-of-distribution questions not covered by existing training data in each iteration. 

Therefore, we argue that \emph{practical \Problem models should be built with strong generalizability to out-of-distribution questions at test time}. 
More specifically, we propose three levels of generalization: \emph{\iid}, \emph{compositional}, and \emph{zero-shot} (\autoref{fig:intro}). 
In addition to the standard \iid generalization, \Problem models should also generalize to novel compositions of seen schema items (relations, classes, functions). 
For example, if a model has been trained with questions about relations like \wl{producer}, \wl{staged\_here}, \wl{capacity}, classes like \wl{Theater}, and functions like \wl{GE}, it should be able to answer complex questions involving all these schema items even though this specific composition is not covered in training.
Furthermore, a \Problem model may also encounter questions about schema items or even entire domains that are not covered in training at all (e.g., \wl{TV\_Program} and \wl{program\_created}) and needs to generalize in a zero-shot fashion~\cite{palatucci2009zero}. 

High-quality datasets are of great importance for the community to advance towards \Problem models with stronger generalization. In addition to providing a benchmark for all three levels of generalization, an ideal dataset should also be large-scale, diverse, and capture other practical challenges for \Problem such as entity linking, complex questions, and language variation. However, existing \Problem datasets are usually constrained in one or more dimensions. 
Most of them are primarily focused on the \iid setting~\cite{berant-etal-2013-semantic,yih-etal-2016-value,DBLP:journals/corr/BordesUCW15,talmor-berant-2018-web}. \GraphQ~\cite{su-etal-2016-generating} and \QALD~\cite{qald9} could be used to test compositional generalization but not zero-shot generalization. They are also relatively small in scale and have limited coverage of the KB ontology. \SimpleQ~\cite{DBLP:journals/corr/BordesUCW15} is large in scale but only contains single-relational questions with limited diversity.
% \ComplexQ\cite{talmor-berant-2018-web} is constructed by automatically extending the logical forms in \WebQ\cite{berant-etal-2013-semantic,yih-etal-2016-value} with additional constraints, but its coverage of the KB ontology is still limited and such extension may lead to artificial questions.
A quantitative comparison can be found in Table~\ref{table:dataset}.

Therefore, we construct a new large-scale, high-quality dataset for \Problem, named \OurDataset (Strongly \underline{G}ene\underline{ra}l\underline{i}zab\underline{l}e \underline{Q}uestion \underline{A}nswering), that supports evaluation of all three levels of generalization. It contains 64,331 crowdsourced questions involving up to 4 relations and functions like counting, comparatives, and superlatives, making it the largest \Problem dataset with complex questions to date. The dataset covers all the 86 domains in \Freebase \Commons, the part of \Freebase considered to be high-quality and ready for public use, and most of the classes and relations in those domains. Furthermore, the questions in our dataset involve entities spanning the full spectrum of popularity from \wl{United\_States\_of\_America} to, e.g.,  \wl{Tune\_Hotels} --- a small hotel chain mainly operated in Malaysia. To make the dataset even more diverse and realistic, we collect common surface forms of entities, e.g., \nl{Obama} and \nl{President Obama} for \wl{Barack\_Obama}, via large-scale web mining and crowdsourcing and use them in the questions. Finally, we develop multiple quality control mechanisms for crowdsourcing to ensure dataset quality, and the final dataset is contributed by 6,685 crowd workers with highly diverse demographics. 

In addition to the dataset, we also study important factors towards stronger generalization and propose a novel \Problem model based on pre-trained language models like BERT~\cite{devlin-etal-2019-bert}. We first show that our model performs competitively with existing models: On \GraphQ, our model sets a new state of the art, beating prior models by a good margin (3.5\%). On \OurDataset, our model significantly outperforms a state-of-the-art \Problem model. More importantly, the combination of our dataset and model enables us to thoroughly examine several challenges in \Problem such as search space pruning and language-ontology alignment. The comparison of different variants of our model clearly demonstrates the critical role of BERT in compositional and zero-shot generalization. 
To the best of our knowledge, \textit{this work is among the first to demonstrate the key role of pre-trained contextual embeddings such as BERT at multiple levels of generalization for \Problem}.\nop{ Fine-grained comparison of different models clearly shows their respective strengths and weaknesses as well as the challenges of our dataset, and points out promising venues for further improvement.} Finally, we also show that our dataset could serve as a valuable pre-training corpus for \Problem in general: pre-trained on our dataset, our model can generalize to other datasets (\WebQ) and reach similar performance fine-tuned with only 10\% of the data, and can even generalize in a zero-shot fashion across datasets. To summarize, the key contributions of this paper are three-fold:

\begin{itemize}[topsep=.3em]
    \item We present the first systematic study on three levels of generalization, i.e., \textit{\iid}, \textit{compositional}, and \textit{zero-shot}, for \Problem and discuss their importance for practical \Problem systems.
    
    \item We construct and release to the public a large-scale, high-quality \Problem dataset containing 64K questions with diverse characteristics to support the development and evaluation of \Problem models with stronger generalization at all three levels.
    
    \item We propose a novel BERT-based \Problem model with competitive performance and thoroughly examine and demonstrate the effectiveness of pre-trained contextual embeddings in generalization. We also present fine-grained analyses which point out promising venues for further improvement.  
\end{itemize}

% To summarize, our main contributions are:
% \begin{itemize}
%     \item We release a large-scale KBQA dataset of high quality with annotated logical forms. With the new dataset, we can standardize the research of KBQA and thus facilitate its development.
%     \item We implement several baselines for the new dataset and point a direction for utlizing BERT in KBQA.
% \end{itemize}
\section{Background}

\subsection{Knowledge Base}
\label{sec:knowledge_base}

\begin{figure*}[!th]
\centering
\includegraphics[width=.97\linewidth]{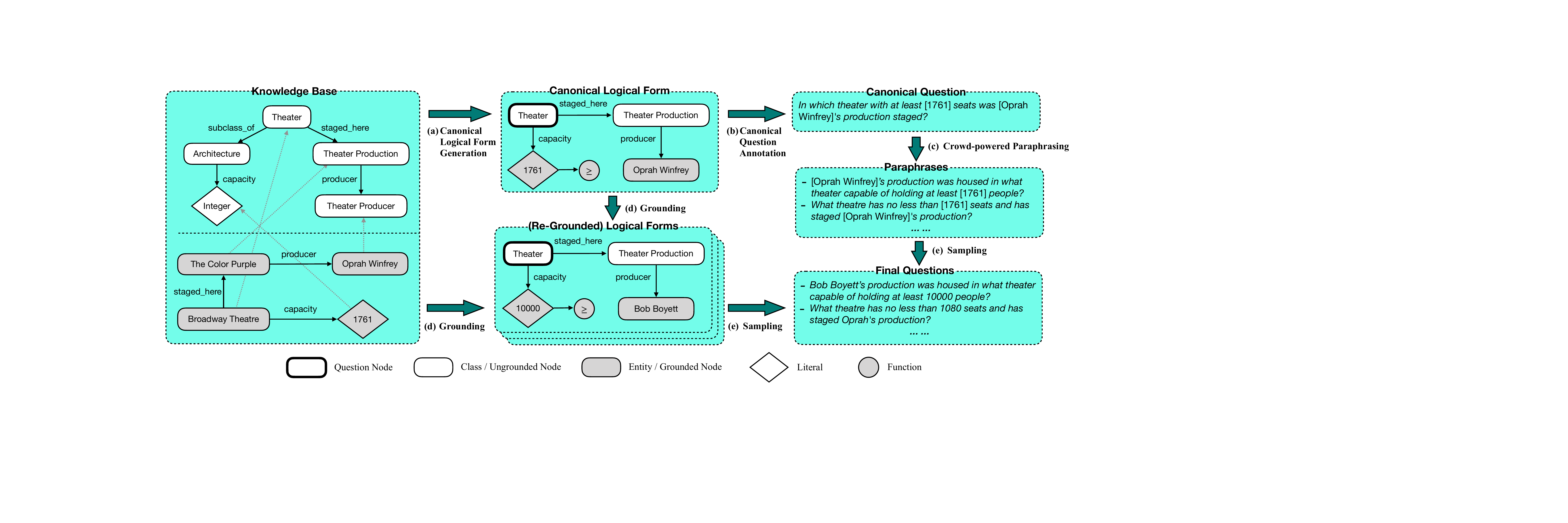}
% \vspace{-10pt}
\caption{Our data collection pipeline with illustration of how one example question in~\autoref{fig:intro} is derived: (a) Generate canonical logical forms from the given KB up to the specified complexity.\nop{ We use the graph-shaped meaning representation and the logical form generation algorithm from \citet{su-etal-2016-generating}.} (b) Experts annotate canonical questions for the canonical logical forms. Entities and literals are enclosed in brackets so that they can be easily replaced. (c) Get high-quality and diverse paraphrases via crowdsourcing. (d) Generate different logical forms from each canonical logical form with different entity groundings. (e) Sample different combinations of logical form and paraphrase to generate the final questions. We additionally mine common surface forms of entities from the web, e.g., \nl{Oprah} for \wl{Oprah\_Winfrey}, and use them in the final questions to make entity linking more realistic.}
\label{fig:data_collection}
\end{figure*}

A knowledge base consists of two parts, an ontology $\mathcal{O}\subseteq\mathcal{C}\times\mathcal{R}\times\mathcal{C}$ and the relational facts $\mathcal{M}\subseteq\mathcal{E}\times\mathcal{R}\times(\mathcal{C}\cup\mathcal{E}\cup\mathcal{L})$, where \Cls is a set of classes, \E is a set of entities, \L is a set of literals and \R is a set of binary relations. 
An example is shown in~\autoref{fig:data_collection}, where the top part is a snippet of the \Freebase ontology and the bottom part is some of the facts.
We base our dataset on the latest version of \Freebase.
Although \Freebase has stopped getting updated, it is still one of the largest publicly available KBs, and its high quality from human curation makes it a solid choice for benchmarking \Problem. We use its \Commons subset which contains 86 domains, 2,038 classes, 6,265 relations, and over 45 million entities.\nop{ and get rid of ad-hoc domains under \wl{base} and \wl{user}. }

\subsection{Three Levels of Generalization: Definition}
\label{sec:generalization_definition}

The definitions are based on the underlying logical form of natural language questions. We denote $\mathcal{S}$ as the full set of \emph{schema items} that includes $\mathcal{R}$, $\mathcal{C}$, and optionally a set of language-specific constructs from the meaning representation language (e.g., SPARQL). In our case we include the set of functions (Section~\ref{sec:data}). Note that entities and literals are not included. Denote $\mathcal{S}_{train}$ as the set of schema items in any of the training examples and $\mathcal{S}_{q}$ that of a question $q$. A qualifying test set $\mathcal{Q}$ for each level of generalization is defined as:

\begin{itemize}
    \item \textbf{\IID generalization}: $\forall q \in \mathcal{Q}, \mathcal{S}_{q} \subset \mathcal{S}_{train}$. In addition, the test questions follow the training distribution, e.g., randomly sampled from the training data.
    
    \item \textbf{Compositional generalization}: $\forall q \in \mathcal{Q}, \mathcal{S}_{q} \subset \mathcal{S}_{train}$, however, the specific logical form of $q$ is not covered in training.
    
    \item \textbf{Zero-shot generalization}: $\forall q \in \mathcal{Q}, \exists s \in \mathcal{S}_{q}, s \in \mathcal{S} \setminus \mathcal{S}_{train}$.
\end{itemize}

The levels of generalization represent the expectation on \Problem models: A model should minimally be able to handle questions \iid to what it has been trained with. One step further, it should also generalize to novel compositions of the seen constructs. Ideally, a model should also handle novel schema items or even entire domains not covered by the limited training data, which also includes compositions of novel constructs. By explicitly laying out the different levels of generalization, we hope to encourage the development of models built with stronger generalization. Orthogonal to these, practical \Problem models should also have strong generalization to language variation, i.e., different paraphrases corresponding to the same logical form. The dataset we will introduce next also puts special emphasis on this dimension.

%Freebase has over 6K relations, 2K classes, 40M entities and 3B facts. A snippet of Freebase is shown in fig.~\ref{fig:run}, where entities are denoted as shaded rectangles, literals are denoted as shaded diamonds and their corresponding classes are represented in white nodes. 

\nop{
\subsection{Meaning Representation} 
We represent logical forms as Lisp programs for ease of composition and readability. The semantics is closely related to lambda DCS~\cite{lambdadcs}. It employs set-based semantics: Each function takes a number of arguments, and both the arguments and the denotation of the functions are either a set of entities or entity tuples. There are 10 functions defined in our Lisp grammar, including \textrelation{COUNT} that returns the cardinality of a set, \textrelation{R} that denotes reverse relations, \textrelation{AND} that denotes set intersection, \textrelation{JOIN} that bridges two sets, superlatives (\textrelation{ARGMAX} and \textrelation{ARGMIN}), and comparatives (\textrelation{LT, GT, LE, GE}). Take Figure~\ref{fig:run} as example. The second \textrelation{JOIN} finds all theater productions produced by \textentity{Bob\_Boyett} and the first \textrelation{JOIN} uses the resulted set as an argument to find all the theaters that have staged any of \textentity{Bob\_Boyett}'s productions. Function \textrelation{GE} returns venues whose capacity is larger than or equal to 10,000. Finally, \textrelation{AND} returns the intersection of the two sets. More detailed definition can be found in Appendix~\ref{sec:lisp}.
}
%our lisp-like logical also include functions like \textrelation{AND, JOIN, COUNT, R, ARGMAX, ARGMIN}, which are similar to lambda DCS. Besides, our logical forms also support 4 comparative functions, i.e., \textrelation{lt, gt, le} and \textrelation{ge}. More details can be found in appendix~\ref{sec:lisp}.
%\subsection{Knowledge Base Question Answering}

\section{Data}
\label{sec:data}

In this section, we describe how our dataset, \OurDataset, is constructed and present analyses on its quality and diversity. How to split the dataset to create evaluation setups for different levels of generalization is described in Section~\ref{sec:exp_setup}.

\subsection{Data Collection}
\label{sec:data_collection}

\begin{table*}[th]
\small
% \begin{tabular}{|p{1cm}|p{1cm}|p{1cm}|p{1cm}|p{1cm}|}
\centering
\resizebox{.8\textwidth}{!}{
\begin{tabular}{lcccccccc}
\toprule
& \textbf{Questions} & \textbf{Canonical LF} & \textbf{Domains} & \textbf{Relations} & \textbf{Classes}  &  \textbf{Entities} & \textbf{Literals} & \textbf{Generalization Assumption} \\ \midrule
\WebQSP\cite{berant-etal-2013-semantic,yih-etal-2016-value}       & 4,737 &  794   & 56      &   661           & 408              &  2,593     & 47 & \iid     \\ \midrule
\ComplexQ\cite{talmor-berant-2018-web} & 34,689  & 6,236  & 61      &  1,181          & 501                   &  11,840    & 996    & \iid  \\ \midrule
\SimpleQ\cite{DBLP:journals/corr/BordesUCW15}     & 108,442 & 2,004 & 76      &  2,004         & 741             &  89,312    & 0    & \iid    \\ \midrule
% \QuAD &30,000& 6,292& N/A & 1,310 & &21,258 &  & \iid\\\midrule
\QALD\cite{qald9} &558  & 558 & N/A & 254 &119 &439 & 28 & comp.\\ \midrule
\GraphQ\cite{su-etal-2016-generating}     & 5,166  & 500  & 70      &  596           & 506           &  376    & 81  & comp.  \\ \midrule
\textbf{\OurDataset}     & 64,331 & 4,969  &  86      &  3,720         & 1,534           & 32,585      & 3,239  & \iid+ comp.+ zero-shot    \\ 
 \bottomrule
\end{tabular}}

\caption{Comparison of \Problem datasets. The number of distinct canonical logical forms (LFs) provides a view into the diversity of logical structures. For example, even though \SimpleQ appears to be the largest, its diversity is limited because it only contains single-relational logical forms with no function (e.g., over 3,000 questions asking about the \wl{place\_of\_birth} of different persons). \nop{It is worth noting that, because our questions are annotated with typed placeholders, it is trivial to generate more questions with different entity groundings. Such data augmentation could be helpful for model robustness and generalization for pre-training (Sec~\ref{sec:transfer}).} Datasets other than \OurDataset are also likely to have test samples including schema items not covered in training, but that is not sufficient to support systematic evaluation of zero-shot generalization. See more details in Appendix~\ref{sec:appendix_comparison}.\nop{ The number of domains for \QALD is N/A because there is no explicit way to get domains from DBpedia.}}
\label{table:dataset}
\vspace{-15pt}
\end{table*}
\begin{table*}[]
    \small
    \centering
    \resizebox{.8\textwidth}{!}{
    \begin{tabular}{lcccc}\toprule
        \multicolumn{1}{c}{\textbf{Question}} & \textbf{Domain} & \textbf{Answer} & \textbf{\# of Relations} &\textbf{Function} \\ \midrule
        %\vtop{\hbox{1.Defender of the crown is a series of games}\hbox{ made for which platform?}}& \\
        \multicolumn{1}{l}{Beats of Rage is a series of games made for which platform?} & Computer Video Game& DOS & 1 & none\\ \midrule
        %2&\begin{tabular}[t]{l}What video games have Minecraft in them?\end{tabular} & cvg & Mining & 1 & none & Y\\ \midrule
        % 2&\multicolumn{1}{l}{\begin{tabular}[t]{l}Which types of storage are compatible with Canon eos 650d \\ as well as color filter array of bayer filter?\end{tabular}} & Digicam & Secure Digital & 3 & none \\ \midrule
        %2&\multicolumn{1}{p{9cm}}{Which types of storage are compatible with Canon eos 650d as well as color filter array of bayer filter?} & Digicam & Secure Digital & 3 & none \\ \midrule
        %2&\multicolumn{1}{p{9cm}}{Where was a film created by director Kasi Lemmons and art director Grant Van Der Slagt?}& \begin{tabular}[t]{c}Location \\ Film\end{tabular} & New York City & 3 & none\\ \midrule
         Which tropical cyclone has affected Palau and part of Hong Kong? & \begin{tabular}[t]{c}Location, \\ Meteorology\end{tabular} & Typhoon Sanba & 3 & none \\ \midrule
        % 3&\multicolumn{1}{l}{What chemical element was first discovered?} & Chemistry & Bismuth & 1 & superlative \\ \midrule
        \multicolumn{1}{l}{Marc Bulger had the most yards rushing in what season?} & \begin{tabular}[t]{c}Sports, \\ American Football\end{tabular} & 2008 NFL Season & 3 & superlative \\ \midrule
        %4&\begin{tabular}[t]{l}Which indonesian province's capital is Gorontalo?\end{tabular} & location & Gorontalo& 1 & none & Y \\ \midrule
        %4&\multicolumn{1}{l}{How many theater plays are in psychological horror?} & Theatre & 4 & 2 & count\\\midrule
        \multicolumn{1}{l}{How many titles from Netflix have the same genre as The Big Hustle?} & Media Common & 20,104 & 2 & count\\\midrule
        \multicolumn{1}{l}{What bipropellant rocket engine has less than 3 chambers?}& Spaceflight & \begin{tabular}[t]{c}RD-114 \\ RD-112, ...\end{tabular} & 1 & comparative \\
         \bottomrule
    \end{tabular}}
    \caption{Example questions from \OurDataset.}
    \label{table:question}
    \vspace{-15pt}
\end{table*}
\citet{wang-etal-2015-building} propose the \Overnight approach which collects question-logical form pairs with three steps: (1) generate logical forms from a KB, (2) convert logical forms into canonical questions, and (3) paraphrase canonical questions into more natural forms via crowdsourcing. \citet{su-etal-2016-generating} extend the \Overnight approach to large KBs like \Freebase with algorithms to generate logical forms that correspond to meaningful questions from the massive space. However, the data collection in \citet{su-etal-2016-generating} was entirely done by a handful of expert annotators, which significantly limits its scalability and question diversity. We build on the approach in \citet{su-etal-2016-generating} and develop a crowdsourcing framework with carefully-designed quality control mechanisms, which is much more scalable and yield more diversified questions. As a result, we are able to construct a dataset one order of magnitude larger with much better coverage of the \Freebase ontology (\autoref{table:dataset}). Our data collection pipeline is illustrated in~\autoref{fig:data_collection}.

\begin{remark}[Canonical logical form generation.] We leverage the logical form generation algorithm from \citet{su-etal-2016-generating}, which randomly generates logical forms with rich characteristics. The algorithm guarantees some important properties of the generated logical forms such as well-formedness and non-redundancy. It first traverses the KB ontology to generate graph-shaped templates that only consist of classes, relations, and functions, and then ground certain nodes to compatible entities to generate logical forms in their meaning representation called \textit{graph query}\footnote{Refer to \citet{su-etal-2016-generating} for the syntax and semantics of graph query.}. At this stage, we only ground each template with one set of compatible entities to generate one \textit{canonical logical form} and use it to drive the subsequent steps. Following prior work, we generate logical forms containing up to 4 relations and optionally containing one function selected from counting, superlatives (\wl{argmax}, \wl{argmin}), and comparatives ($>$, $\geq$, $<$, $\leq$). The exemplar logical form in~\autoref{fig:data_collection} has 3 relations and one function. 

Each canonical logical form is validated by a graduate student against the criterion -- \emph{could we reasonably expect a real human user to ask this question?} We only keep the valid ones, which reduces the number of artificial questions that are common in other KBQA datasets with generated logical forms such as \ComplexQ~\cite{talmor-berant-2018-web} and \SimpleQ~\cite{DBLP:journals/corr/BordesUCW15} and makes the dataset more realistic. The overall approval rate at this step is 86.5\%.
\end{remark} 

\begin{remark}[Canonical question annotation.]
Each validated canonical logical form is annotated with a canonical question by a graduate student, which is then cross-validated by another student to ensure its fidelity and fluency. The graduate students are all knowledgeable of \Problem and are trained with detailed materials about the annotation task. To further facilitate the task, we develop a graphical interface (\autoref{fig:annotation}) which displays an interactive graphical visualization of the logical form. One can click on each component to see auxiliary information such as a short description and its domain/class. Annotators are required to enclose entities and literals in brackets (\autoref{fig:data_collection}).
\end{remark}

\begin{remark}[Crowd-powered paraphrasing.] We use Amazon Mechanical Turk to crowdsource paraphrases of the canonical questions\nop{\footnote{See distribution of paraphrases in Appendix~\ref{sec:appendix_dist}.}} and limit our task to native English speakers with at least 95\% task approval rate. We develop a crowdsourcing framework with automated quality control mechanisms that contains three tasks (instructions in~\autoref{fig:crowdsourcing_instructions}):

\begin{itemize}[itemsep=.2em,topsep=.2em]
    \item \emph{Task 1: Paraphrasing}. A crowd worker is presented with a canonical question as well as auxiliary information such as topic entity description and answer to the question, and is tasked to come up with a plausible and natural paraphrase. Entities and literals enclosed with brackets are retained verbatim in the paraphrase. The system keeps running until 5 \textit{valid} paraphrases (Task 2) are obtained for each canonical question. We pay \$0.10 per paraphrase.
    
    \item \emph{Task 2: Cross-validation}. Each paraphrase is then judged by multiple independent workers (the original author is excluded) on its fluency and fidelity (i.e., semantic equivalence) to the corresponding canonical question. Each paraphrase gets 4 judgements on average and those with fidelity approval rate below 75\% \emph{or} fluency approval rate below 60\% are discarded. Overall 17.4\% of paraphrases are discarded. A worker judges 10 paraphrases a time, one of which is a \textit{control question}. We manually craft an initial set of paraphrase-canonical question pairs known to be good/bad as control questions and gradually expand the set with new pairs validated by crowd workers. We ask workers to reconsider their validation decision of the whole batch when they have failed the latent control question, and we found that by doing so spammers or low-quality workers tend to quit the task and we end up with mostly high-quality workers. We pay \$0.15 per batch. \nop{Workers with a high disapproval rate are banned from the task but still get paid for what they have done.} 
    
    \item \emph{Task 3: Entity surface form mining}. We collect a list of common surface forms ranked by frequency for each entity from a large entity linking dataset, FACC1~\cite{gabrilovich2013facc1}, which identifies around 10 billion mentions of \Freebase entities in over 1 billion web documents from ClueWeb. For example, some common surface forms (and frequency) for \wl{Barack\_Obama} are \nl{obama} (21M), \nl{barack obama} (5.3M), and \nl{barack hussein obama} (101K). We then ask at least three crowd workers to select true surface forms from the mined list, and those selected by less than 60\% of the workers are discarded. We pay \$0.05 per task.  
\end{itemize}

\end{remark}

\begin{remark}[Grounding and sampling.]
More logical forms are generated by grounding each canonical logical form with \textit{compatible} entity groundings.\footnote{An entity grounding, e.g., $($\wl{Bob Boyett}, \wl{10000}$)$, is compatible with a canonical logical form if the re-grounded logical form yields a non-empty denotation on the KB.} 
We do controlled sampling to generate the final questions: From the pool of logical forms and paraphrases associated with the same canonical logical form, we sample one from each pool at a time to generate a question (\autoref{fig:data_collection}). 
We start with uniform weights and each time a logical form or paraphrase is selected, its weight is divided by $\rho_l$ and $\rho_p$, respectively. 
We set $\rho_l$ to 2 and $\rho_p$ to 10 to enforce more linguistic diversity. Finally, we randomly replace entity surface forms with the ones mined in Task 3 (if there is any). 
This way, we are able to generate a large-scale dataset with carefully-controlled quality and diversity.
\end{remark}

\subsection{Dataset Analysis}

In total we have collected 4,969 canonical logical forms and 29,457 paraphrases (including the canonical questions). The final dataset contains 64,331 question-logical form pairs after sampling, which is by far the largest dataset on \Problem with complex questions. We note that such sampling is a unique feature of \OurDataset because of our data collection design.
A detailed comparison with existing datasets is shown in~\autoref{table:dataset} and some example questions are shown in~\autoref{table:question}. 
\OurDataset has significantly broader coverage and more unique canonical logical forms than existing dataset except \ComplexQ.\footnote{Logical forms in \ComplexQ are automatically extended from \WebQ with extra SPARQL statements. This adds structural diversity but may lead to unnatural questions.} 
\OurDataset is also the only \Problem dataset that explicitly evaluates zero-shot generalization (Section~\ref{sec:exp_setup}). 
Question distributions over different characteristics are shown below:

\begin{table}[!h]
    \small
    \centering
    
    \resizebox{\linewidth}{!}{
    \begin{tabular}{cccc|cccc|cc}
    \toprule
    \multicolumn{4}{c}{\textbf{\# of Relations}} & \multicolumn{4}{c}{\textbf{Function}} & \multicolumn{2}{c}{\textbf{Answer Cardinality}} \nop{& \multicolumn{2}{c}{\textbf{w/ Entity?}}} \\ \midrule
        1 & 2 & 3 & \multicolumn{1}{c|}{4} & none & count & super. & \multicolumn{1}{c|}{comp.} & 1 & \multicolumn{1}{c}{$>$ 1} \nop{& Y & N }\\
    \midrule
        % \nop{\# of graph queries & 3,454 & 1,284 & 242 & 10 & 3,877 & 239 & 603 & 271 & 3,434 & 1,556 & 3,840 & 1,150\\ \midrule}
        % \# of questions & 32,663 & 12,406 & 2,383 & 100 & 38,723 & 2,462 & 3,998 & 2,469 & 32,547 & 15,005 & 38,599 & 8,953\\
          44,340 & 16,610 & 3,254 & 127 & 53,526 & 3,463 & 4,111 & 3,231 & 44,044 & 20,287 \nop{& 53,705 & 10,911}\\
    \bottomrule
    \end{tabular}}
\end{table}

\begin{remark}[Quality.]
As a qualitative study on data quality, we manually analyze 100 randomly sampled canonical logical forms, their canonical questions, and the associated paraphrases: 3 of the 100 canonical questions have mismatching meaning with the canonical logical form, mostly due to misinterpreting the directionality of some relation, and 12 of the 576 paraphrases do not match the corresponding canonical question, leading to 3\% error rate of canonical question annotation, 2.1\% error rate of paraphrasing, and 5.6\% error rate overall. All the examined paraphrases are reasonably fluent.  
\end{remark}

\begin{remark}[Linguistic diversity.]
The dataset is contributed by 11 graduate students and 6,685 crowd workers with diverse demographics in terms of age group, education background, and gender (Appendix~\ref{sec:appendix_demographics}). A common concern with crowd-powered paraphrasing as a means of data collection is lack of linguistic diversity --- crowd workers may be biased towards the canonical question~\cite{herzig2019don}. However, that may not be the case for \OurDataset. The average Jaccard similarity between each paraphrase and the corresponding canonical question, when all lower-cased, is 0.569 and 0.286 for unigrams and bigrams, respectively. The average Levenshtein edit distance is 26.1 (with average total length of 50.1 characters). The same statistics between the paraphrases of the same canonical question are 0.527, 0.257, and 28.2. The dissimilarity is even higher when excluding entities. We believe this is a decent level of linguistic diversity and we attribute it to our diverse crowd workers and quality control mechanisms.

\end{remark}

\begin{remark}[Entity linking.]
\OurDataset is also featured with more realistic and challenging entity linking thanks to our large-scale mining of entity surface forms. Common types of surface forms include acronym (\nl{FDA} for \wl{Food\_and\_Drug\_Administration}), last/first name (\nl{Obama} for \wl{Barack\_Obama}), commonsense (\nl{Her Majesty the Queen} for \wl{Elizabeth\_II}), and colloquial vs.\ formal (\nl{Obama vs.\ Romney} for  \wl{United\_States\_Presidential\_Election\_2012}). In general the mined surface forms are the more typical, colloquial way of referring to an entity than its formal name in the KB, which makes the questions more realistic. We note that this is an important challenge towards practical \Problem systems but is largely neglected in existing \Problem datasets. For example, \citet{yao-2015-lean} shows that, for the \WebQ dataset, simple fuzzy string matching is sufficient for named entity recognition (NER) due to the way the dataset is constructed.
\end{remark}

\subsection{Logical Form in S-expression}
In addition to graph query, we provide an alternative linearized version of it in S-expressions, which is needed to apply the mainstream sequence-to-sequence (Seq2Seq) neural models~\cite{jia-liang-2016-data, dong-lapata-2016-language, zhang-etal-2019-complex}. We find that S-expression provides a good trade-off on compactness, compositionality, and readability, but one may choose any other formalism for linearization such as $\lambda$-DCS~\cite{liang2013lambda} or directly use the corresponding SPARQL queries. Every graph query can be easily converted to an equivalent S-expression. For example, The graph query in~\autoref{fig:data_collection}(d) can be converted into \textrelation{(AND Theater (AND (GE capacity 10000) (JOIN staged\_here (JOIN producer Bob\_Boyett))))}. More details can be found in Appendix~\ref{sec:lisp}. Both graph queries and S-expressions can be easily converted into SPARQL queries to get answers. We will use S-expressions as logical form in our modeling.
\section{Modeling}
\label{sec:model}

In this section, we discuss some of the challenges arising from non-\iid generalization and potential solutions. The proposed models, combined with \OurDataset, will enable us to evaluate and compare different strategies for stronger generalization. 

Compositional and zero-shot generalization pose two unique challenges compared with \iid generalization: \textit{large search space} and \textit{language-ontology alignment}. The first challenge is the significantly larger search space. Under the \iid assumption, a model only needs to consider the part of the ontology observed during training. For zero-shot generalization, however, one could not assume that to hold at test time and needs to consider the entire ontology. Models that build their vocabulary solely from training data (e.g., \cite{zhang-etal-2019-complex}) would almost certainly fail at zero-shot generalization. \emph{Question-specific search space pruning} is thus more important than it is in the \iid setting. Secondly, a major challenge in \Problem is to build a precise alignment between natural language and schema items in the ontology in order to understand what each question is asking about. While it is already important for \iid generalization, it becomes even more critical for non-\iid generalization: For zero-shot generalization, one apparently needs to understand the unseen schema items in order for it to possibly handle the corresponding questions. Even for compositional generalization, where the schema items are covered in training, precise language-ontology alignment is still critical in order for the model to generate novel compositions other than something it has memorized in training. Conventional methods~\cite{berant-etal-2013-semantic, Abujabal, cai-yates-2013-large} use web mining to build a lexicon from natural language phrases to KB schema items, which may be biased towards popular schema items. The emergence of pre-trained contextual embeddings such as BERT~\cite{devlin-etal-2019-bert} provides an alternative solution to building language-ontology alignment. Next, we propose a BERT-based \Problem model which will enable us to examine the effectiveness of pre-trained contextual embeddings in non-\iid generalization. 
\begin{figure*}[th]
\centering
\includegraphics[width=0.8\linewidth]{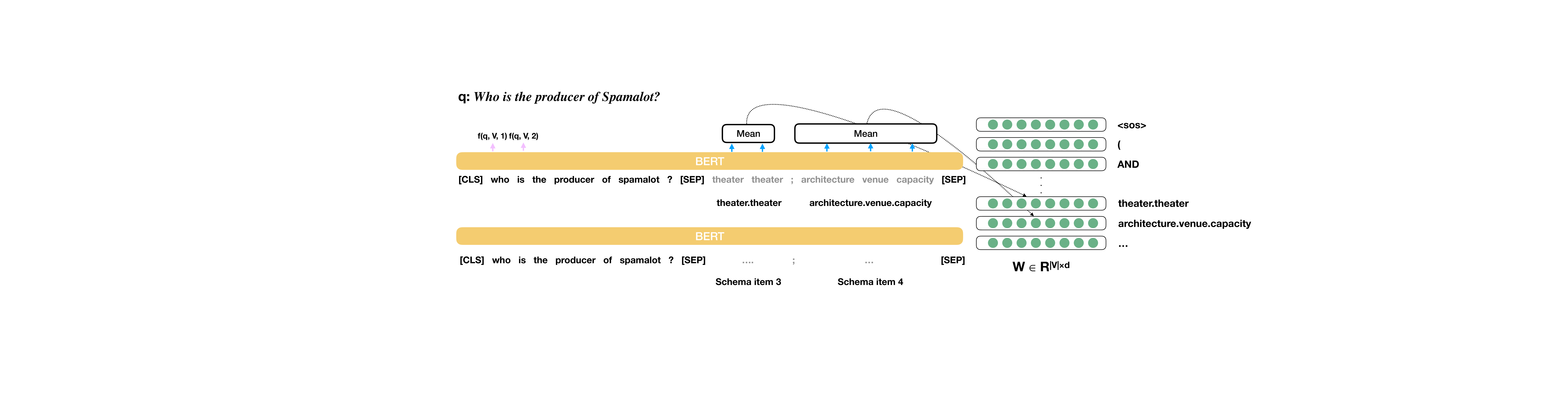}
\caption{An overview of using BERT to jointly encode the input question and schema items in the decoder vocabulary. We evenly split all items in the vocabulary into chunks and concatenate each chunk with the question one by one. For the purpose of visualization, here we set the chunk size as 2. The first chunk contains two schema items, \textrelation{theater.theater} and \textrelation{architecture.venue.capacity}. For each item in $\mathcal{V}$, we compute its embedding in $\textbf{W}$ by averaging the last layer's output of BERT of all its word-pieces. To compute $f(q, \mathcal{V}, t)$, we simply take the last layer's output of BERT in the first chunk for the $t$-th word-piece of the question. Note that, we show whole words instead of actual word-pieces in the figure for brevity.}
\label{fig:bert}
\vspace{-10pt}
\end{figure*}
\nop{
In the deep learning era, such language knowledge can be found in pre-trained word embeddings or language models. Among others, one of the most popular pre-trained language models, BERT~\cite{devlin-etal-2019-bert}, has achieved great success in learning an alignment for unseen items in cross-domain text-to-SQL tasks.\nop{ Also, as pointed in~\cite{arora2020contextual}, BERT shows significantly better performance than uncontexualized word embeddings in zero-shot setting.} Nonethelss, uncontextualized embeddings suffice to achieve good performance on previous \iid \Problem datasets, and as a result, pre-trained language models are still being under-exploited in \Problem. To better address the two unique challenges and to fill the gap in the literature, we propose a new model that takes advantage of BERT to learn better alignment for unseen items and uses entities as anchor to effectively prune the search space. 
}

\nop{
To handle the enormous search space, it is critical to use entities identified from the question as anchor and base on the anchor's context in KB to prune the search space. One common practice is generating candidate logical forms from the anchor entities following some pre-specified templates or rules.}

\subsection{Model Overview}
\label{sec:parser}
Our goal is to learn a model that maps an input question $q = x_1, ..., x_{|q|}$ to its corresponding logical form $a = y_1, ..., y_{|a|}$.
The entire model is based on Seq2Seq~\cite{Seq2Seq, DBLP:journals/corr/BahdanauCB14}, which is commonly used in many semantic parsing tasks~\cite{dong-lapata-2016-language, jia-liang-2016-data,hwang2019comprehensive, guo-etal-2019-towards, zhang-etal-2019-editing}. Seq2Seq comprises an encoder that encodes input $q$ into vectors and a decoder that autoregressively output one token $y'\in \mathcal{V}$ conditioned on the input representation, where $\mathcal{V}$ is the decoding vocabulary. Specifically, the conditional probability $p(a|q)$ is decomposed as:

\begin{equation}
    p(a|q) = \prod_{t=1}^{|a|}p(y_t|y_{<t}, q),
\end{equation}

\noindent where $y_{<t} = y_1,...,y_{t-1}$.
Our encoder and decoder are two different recurrent neural networks with long short-term memory units (LSTM)~\cite{hochreiter1997long}. LSTM recursively processes one token at each time step as the following function:

\begin{equation}
    \textbf{h}_t = LSTM(\textbf{h}_{t-1}, \textbf{g}_t),
\end{equation}

\noindent where $\textbf{g}_t$ is the representation of the input token at the $t$-th step, and $\textbf{h}_{t-1}$ is the hidden state from last time step. During encoding, $\textbf{g}_t = f(q, \mathcal{V}, t)$, with $f$ being a function takes $q, \mathcal{V}, t$ as input and returns the embedding for $x_t$; during decoding, $\textbf{g}_t = [\textbf{W}]_{y_t}$, which denotes the the embedding corresponding to $y_t$ in the embedding matrix $\textbf{W}\in \textbf{R}^{|\mathcal{V}|\times d}$. We will elaborate how we use BERT to define $f$ and $\textbf{W}$ in the next subsection.

During decoding, the model computes the probability for each output token based on the current hidden state $\textbf{h}_t$ and $\textbf{W}$:

\begin{equation}
    p(y_t|y_{<t}, q) = [Softmax(\textbf{W}\textbf{h}_t)]_{y_t}.
\end{equation}

This means we tie the input and output embeddings using $\textbf{W}$ in Seq2Seq terminology. In this way, we can assign semantic
meanings to words from pre-trained embeddings, which can facilitate the open vocabulary learning~\cite{venugopalan2017captioning}.

\subsection{BERT Encoding}
Now we discuss how we use BERT to compute $\textbf{W}$ and $f(q, \mathcal{V}, t)$. BERT has demonstrated its effectiveness in learning good alignment in cross-domain text-to-SQL parsing with its contextual representation. Specifically, in text-to-SQL, \citet{hwang2019comprehensive} successfully apply BERT by concatenating the question and all vocabulary items together to construct the input for BERT. However, the vocabulary in \Problem is much larger. Directly doing this will exceed BERT's maximum input length of 512 tokens. To address this, we split items from $\mathcal{V}$ into chunks and then concatenate each chunk independently with $q$, and feed them to BERT one by one. Figure~\ref{fig:bert} depicts an example where each chunk has 2 items from $\mathcal{V}$. Specifically, $q$ is separated with each chunk by a special token "[SEP]", and the items inside the same chunk are delimited by ";". Then $f(q, \mathcal{V}, t)$ can be simply defined as the $t$-th output from BERT for the first chunk. For $\textbf{W}$, each row in it is computed by taking average over BERT's output for the constituting words of the corresponding item. For example, the row corresponding to \textrelation{architecture.venue.capacity} in $\textbf{W}$ is computed by averaging the output from BERT for word \nl{architecture}, \nl{venue} and \nl{capacity} in the first chunk as shown in Figure~\ref{fig:bert}.

\begin{remark}[Vocabulary construction.]
We have not talked about what constitutes $\mathcal{V}$. The most trivial way is just to include all schema items from KB ontology to deal with zero-shot generalization. However, this will not only lead to an enormous search space during decoding that makes the prediction very difficult, but also makes it extremely time-consuming for training as we need to create a large number of chunks to feed to BERT for every single question. To reduce the size of $\mathcal{V}$, we leverage entities identified from $q$ as anchors in the KB, and choose to only include KB items that are reachable by at least one of the anchor entities within 2 hops in KB. We refer to this as vocabulary pruning (VP), which is applied during both training and inference. This means we have a dynamic $\mathcal{V}$ for different input $q$. Note that all the identified entities, functions, and syntax constants used in S-expressions are also included. We similarly get the representation for an entity based on the output from BERT that corresponds to its surface form.
\end{remark}

\subsection{Entity Linking}
\label{sec:entity_linker}

Entity linking in most of the existing \Problem datasets is not a major challenge, and exhaustive fuzzy string matching~\cite{yao-2015-lean} may suffice to achieve a reasonable performance. However, the entities in \OurDataset span the full spectrum of popularity and appear in surface forms mined from the web, which is more realistic but also makes entity linking more challenging. 
We use a BERT-based NER system\footnote{https://github.com/kamalkraj/BERT-NER} and train it on our training set. For entity disambiguation from the identified mentions, we simply select the most popular one(s) based on FACC1. As a comparison, we also try Aqqu~\cite{aqqu}, a rule-based entity linker using linguistic and entity popularity features and achieves high accuracy on \WebQSP. The entity linking results on \OurDataset is shown as follows:
\begin{table}[H]
    \small
    \centering
    \begin{tabular}{lccc}
    \toprule
        & \textbf{Recall} & \textbf{Precision} & \textbf{F1} \\\midrule
        \textbf{Aqqu} &  \textbf{77.8} & 9.7 & 17.2\\
        \textbf{BERT} &  77.0 & \textbf{68.0} & \textbf{72.2}\\
    \bottomrule
    \end{tabular}
\end{table}
\noindent The BERT-based entity linker is slightly worse in recall but much better in precision. We will therefore use the BERT-based entity linker afterwards.\nop{ Sometimes multiple entities can be identified in a single question. In that case, we separately do VP (candidate logical forms generation) for each entity and take the union of them for Seq2Seq+VP (\Ranking). }

\subsection{Inference}
We propose two different modes of doing inference using our model, namely \Transduction and \Ranking. For \Transduction, we simply use the model in the normal way, i.e., use the model to autoregressively predict one token from $\mathcal{V}$ at each time step until the model predicts the stop token. For \Ranking, instead of using Seq2Seq as a generator we use it as a ranker to score each candidate logical form and return the top-ranked candidate. We employ a simple yet effective strategy to generate candidate logical forms: We enumerate all logical forms, optionally with a \wl{count} function, within 2 hops starting from each entity identified in the question.\footnote{There are prohibitively many candidates with 3 hops.} The recall is 80\% on \OurDataset. Questions with superlatives and comparatives often do not have a topic entity and are therefore not covered. \Ranking can prune the search space more effectively than the first one, while \Transduction is more flexible and can handle more types of questions.
% \begin{remark}[Search space pruning.]
% One key challenge for higher-level generalization is that the model may become biased to the search space that is explored during training. We devise two simple strategies to remedy for this. The first strategy is called vocabulary pruning (VP). We propose to prune the search space by using the identified entities in the question as anchors, and limiting the decoder vocabulary to only include schema items that are close to the anchors. Specifically, we prune the vocabulary by excluding schema items that cannot be reached from any of the anchor entities within 2 hops in the KB. This can be achieved by using a vocabulary mask for each question, similar to~\cite{yu-etal-2018-spider}. Secondly, we train a Seq2Seq model as before, but instead of using it as a generator we use it as a ranker and return the top-ranked candidate. We employ a simple yet effective strategy to generate candidate logical forms: We enumerate all logical forms, optionally with a \textrelation{count} function, with within 2 hops\footnote{There are prohibitively many candidates with 3 hops.} starting from each entity identified in the question. The recall is 80\% on \OurDataset. Questions with superlatives and comparatives often do not have a topic entity and are therefore not covered. The second strategy can prune the search space more effectively than the first one, while the first strategy is more flexible and can handle more types of questions.
% \end{remark}

\nop{\subsection{Entity Linking}
\label{sec:entity_linking}
Entity linking in most of the existing OpenSP datasets is not a major challenge. For example, because of the way \WebQ was constructed, the topic entities are biased towards the common ones~\cite{su-etal-2016-generating}.
}

\begin{remark}[Effectiveness on existing dataset.]
\nop{Most existing methods either adopt specialized design under \iid assumption (e.g.,\cite{zhang-etal-2019-complex}) or can be hard to adapt to new dataset without source code available. Therefore, we will mainly rely on our new model to study the challenges in our new dataset.} 
Before getting to experiments on our \OurDataset dataset, we conduct an experiment on the existing dataset \GraphQ to show the competitive performance of our model. Our \Ranking model achieves an F1 of 25.0\%, which significantly outperforms the prior art SPARQA~\cite{sun2020sparqa} by 3.5 percent. With this superior performance, we believe our new model is reasonable to serve as a strong baseline on \OurDataset and support the subsequent investigations on three levels of generalization.
% as shown in the following table:
% \begin{table}[H]
%     \small
%     \centering
%     \begin{tabular}{cc}
%     \toprule
%         & \textbf{F1} \\
%     \midrule
%         PARASEMPRE & 12.8 \\ 
%         SCANNER & 17.0 \\
%         UDEPLAMBDA & 17.7 \\
%         PARA4QA & 20.4 \\
%         SPARQA & 21.5 \\
%     \midrule
%         \Transduction & 15.0 \\
%         \Ranking & \textbf{25.0} \\
%     \bottomrule
%     \end{tabular}
% \end{table}
\end{remark}

\nop{\begin{remark}[Seq2Seq.] 
The Seq2Seq model~\cite{Seq2Seq, DBLP:journals/corr/BahdanauCB14} is commonly used in conventional semantic parsing~\cite{dong-lapata-2016-language, jia-liang-2016-data} \nop{It performs semantic parsing in a sequence \Transduction style, where an encoder first encodes the question into a low-dimensional vector, and then a decoder makes sequential predictions (e.g., a sequence of tokens) conditioned on the encoder's output to compose logical forms.}and is a default choice of base architecture for text-to-SQL parsing~\cite{hwang2019comprehensive, guo-etal-2019-towards, zhang-etal-2019-editing}. However, it is not as commonly used for OpenSP. We include Seq2Seq as a baseline because its well-understood properties make it a good reference for comparison. Specifically, we use an attentive Seq2Seq model that encodes a question and decodes a linearized Lisp program. Both encoder and decoder use LSTM~\cite{hochreiter1997long} and the attention is dot product. The decoder vocabulary includes all the classes and relations in the KB, as well as the identified entities in the question and Lisp functions. We use pre-trained GloVe~\cite{pennington-etal-2014-glove} embedding to embed both input and output tokens, which is necessary for zero-shot cross-domain generalization. For multi-word output tokens like KB relations, embeddings are computed by averaging over all the words in the surface form.

\nop{
to our scenario. We use LSTM for both the encoder and decoder. At each step of decoding, the decoder decides to generate a token from the set of syntactic symbols (i.e., brackets and function names), schema items (i.e., all relations and classes in \Freebase) or entities and values identified in the input question. GloVe~\cite{pennington-etal-2014-glove} is used to get the embeddings for all tokens. The embedding for each schema item and entity is computed by averaging the embeddings for all words in its surface form. }
\end{remark}}

\nop{\begin{remark}[Seq2Seq + VP.]
Different from text-to-SQL where an input database only comprises dozens of schema items, a unique challenge for OpenSP is the large search space during decoding. 
%The vanilla Seq2Seq model decodes logical forms from a global vocabulary consisting of thousands of schema items, which can be sub-optimal. 
We propose to prune the search space by using the identified entities in the question as anchors, and limiting the decoder vocabulary to only include schema items that are close to the anchors. We call this component vocabulary pruning (VP). Specifically, we prune the vocabulary by excluding schema items that cannot be reached from any of the anchor entities within 2 hops in the KB. In Seq2Seq+VP, each question has a dynamic decoder vocabulary instead of the global vocabulary used in the vanilla Seq2Seq model above. This can be achieved by using a vocabulary mask for each question, similar to~\cite{yu-etal-2018-spider}.
\end{remark}}

\nop{\begin{remark}[\Ranking.]

Instead of Seq2Seq, OpenSP models more often adopt a candidate generation + \Ranking scheme~\cite{berant-etal-2013-semantic,yih-etal-2015-semantic, kbqa, reddy-etal-2017-universal, Abujabal, luo-etal-2018-knowledge, sun2020sparqa}\nop{bao-etal-2016-constraint, hu2018}, where \Ranking is often based on hand-engineered features.\nop{hand-crafted rules or templates are always used in the generation stage, while syntactic parses and hand-crafted features are commonly used in matching.} Here we train a Seq2Seq model as before. But instead of using it as a generator we use it as a ranker and return the top-ranked candidate. We employ a simple yet effective strategy to generate candidate logical forms: We enumerate all logical forms, optionally with a \textrelation{count} function, with within 2 hops\footnote{There are prohibitively many candidates with 3 hops.} starting from each entity identified in the question. The recall is 80\% on \OurDataset. Questions with superlatives and comparatives often do not have a topic entity and are therefore not covered. 
\end{remark}}
\nop{\begin{remark}[BERT.]
Pre-trained contextual embeddings like BERT have shown their remarkable efficacy in cross-domain text-to-SQL parsing~\cite{hwang2019comprehensive, guo-etal-2019-towards, zhang-etal-2019-editing}, but it has not been applied to OpenSP, where the large schema brings unique challenges. We propose to use BERT in place of GloVe to improve cross-domain generalization of OpenSP. For text-to-SQL parsing, all schema items (table headers) are first concatenated with the input utterance, and BERT is then applied on top of that to get contextualized encoding for the utterance and schema items and learning implicit alignment between the two. However, for OpenSP, because of the large schema, doing something like this during fine-tuning is both very costly, if not infeasible, and exceeding BERT's length limit of 512 subwords. Therefore, for fine-tuning we use a dynamic subset of the global schema for each question.\footnote{We use the schema items that meet the following requirements: (1) falling into the same domain(s) as the question, and (2) can be reached within 2 hops from any of the topic entities.} 
During inference, we can still use the full schema since the forward pass consumes less memory. To avoid exceeding BERT's length limit, we divide the schema items into chunks of size 45 and concatenate each chunk with the question to be encoded by the fine-tuned BERT one by one. The representation for each schema item is computed by averaging the last layer of BERT's output for all subwords in it. All other parts remain the same as GloVe-based models. Refer to Appendix~\ref{sec:appendix_architecture} for illustration of the model architecture.
\end{remark}}

\section{Experiments}
\label{sec:experiments}

\begin{table*}[th]
\small
\begin{tabular*}{0.6\textwidth}{l@{\extracolsep{\fill}}cc|cc|cc|cc}
\toprule
& \multicolumn{2}{c}{\textbf{Overall}}   &\multicolumn{2}{c}{\textbf{\IID}}  &\multicolumn{2}{c}{\textbf{Compositional}}    &\multicolumn{2}{c}{\textbf{Zero-shot}}            \\ \hline
%  & \multicolumn{2}{c}{\textbf{Perfect Entity Linking}} & \multicolumn{2}{c}{\textbf{BERT Entity Linker}}               \\ \hline
% & \textbf{Exact Match} & \textbf{F1} & \\ 
\multicolumn{1}{l|}{} & \textbf{EM} & \textbf{F1} & \textbf{EM} & \textbf{F1} & \textbf{EM} & \textbf{F1} & \textbf{EM} & \textbf{F1} \\ 
\midrule
\multicolumn{1}{l|}{\QGG~\cite{lan-jiang-2020-query}} & $-$    & 36.7  & $-$ & 40.5  & $-$   & 33.0  & $-$ & 36.6    \\
\midrule
\midrule
\multicolumn{1}{l|}{\Transduction} & 33.3  & 36.8  & 51.8  & 53.9      & 31.0  & 36.0      & 25.7  & 29.3    \\
\multicolumn{1}{r|}{$-$ VP}                 & 26.6  & 29.6  &  40.1 &  42.7     & 25.6  &   30.0    & 20.6  & 23.4    \\ 
\multicolumn{1}{r|}{$-$ BERT}               & 17.6  & 18.4  & 50.5  & 51.6  & 16.4  & 18.5  & 3.0   & 3.1   \\ 
\multicolumn{1}{r|}{$-$ VP $-$ BERT}        &  15.4 & 16.1  &  48.3 & 49.1  & 13.5  &15.3   & 1.0   & 1.3   \\

\midrule
\multicolumn{1}{l|}{\Ranking}      & \textbf{50.6}  &  \textbf{58.0}     & 59.9  & 67.0       & \textbf{45.5}  &   \textbf{53.9}    & \textbf{48.6}  &  \textbf{55.7}  \\
\multicolumn{1}{r|}{$-$ BERT}               & 39.5  & 45.1  & \textbf{62.2}  & \textbf{67.3}  & 40.0  & 47.8  & 28.9  & 33.8  \\

\bottomrule
\end{tabular*}
\caption{Overall results. ``$-$ VP'' denotes without vocabulary pruning. ``$-$ BERT'' denotes using GloVe embeddings instead of BERT.}
\label{table:overall}
\vspace{-15pt}
\end{table*}

\subsection{Experimental Setup}
\label{sec:exp_setup}

\begin{remark}[Data split.]
We split \OurDataset to set up evaluation for all three levels of generalization. Specifically, our training/validation/test sets contain about 70\%/10\%/20\% of the data, which correspond to 44,337, 6,763 and 13,231 questions, respectively. For the validation and test sets, 50\% of the questions are from held-out domains not covered in training (\textbf{zero-shot}), 25\% of the questions correspond to canonical logical forms not covered in training (\textbf{compositional}), and the rest 25\% are randomly sampled from training (\textbf{\iid}). The \iid and compositional subsets have the additional constraint that the involved schema items are all covered in training. For the zero-shot subset, 5 domains are held out for validation and 10 for test. 
\end{remark}

\begin{remark}[Evaluation metrics.]
We use two standard evaluation metrics. The first one is \textit{exact match accuracy} (EM), i.e., the percentage of questions where the predicted and gold logical forms are semantically equivalent. To determine semantic equivalence, we first convert S-expressions back to graph queries and then determine graph isomorphism.\footnote{\url{https://networkx.github.io}} Unlike string-based or set-based EM~\cite{yu-etal-2018-spider}, our graph isomorphism based EM is both sound and complete. In addition, we also report the F1 score based on the predicted and gold answer sets~\cite{berant-etal-2013-semantic}. This is better suited for models that directly predict the final answers without any intermediate meaning representation and gives partial credits to imperfect answers. We host a local SPARQL endpoint via Virtuoso to compute answers. 
\end{remark}

\subsection{Models}

Our primary goal of the experiments is to thoroughly examine the challenges of different levels of generalization and explore potential solutions. Therefore, we will evaluate an array of variants of our models with different strategies for search space pruning and language-ontology alignment. To better situate our models in the literature and demonstrate their competitive performance, we also adapt \QGG~\cite{lan-jiang-2020-query}, the state-of-the-art model on \ComplexQ and \WebQSP, to \OurDataset. We have also looked into models developed based on \QALD or \QuAD but the adaptation cost would be too high because most of the source codes are not available.
All models use the same entity linker (Section~\ref{sec:entity_linker}).

\begin{remark}[Our model variants.]
For both \Transduction and \Ranking, we introduce a variant that uses GloVe embeddings\footnote{We choose to use GloVe-6B-300d.} instead of BERT for language-ontology alignment. Specifically, we use the average GloVe embedding for each schema item instead of the encodings from BERT. Unlike BERT, GloVe embeddings are not contextual and are thus not jointly encoded with the current natural language question.
This variant will then allow us to examine the role of contextual embedding in generalization. For \Transduction, we additionally introduce a variant that does not employ entity-based vocabulary pruning (VP). This is not applicable to \Ranking because \Ranking always relies on the identified entities for candidate generation. 
We use uncased BERT-base and fine-tune BERT but fix GloVe (similar to previous work~\cite{guo-etal-2019-towards}), because we find fine-tuning GloVe makes it slightly worse.
Other implementation details such as hyper-parameters and training scheme can be found in Appendix~\ref{sec:appendix_setup}.

\nop{
Specifically, for \Transduction, we introduce a variant in which vocabulary pruning based on anchor entities is not employed. We also introduce a variant where we use GloVe embeddings to get the uncontexutalized representation for all items in the decoding vocabulary. Finally, we also compare with \Transduction for which both vocabulary pruning and BERT are discarded. For \Ranking, because \Ranking model does not rely on vocabulary pruning to reduce the search space, which makes the comparison with \Ranking without vocabulary pruning less meaningful, so we only include a variant of it that uses GloVe instead of BERT in our experiments. We fine-tune BERT and fix GloVe, because we find fine-tuning GloVe makes it worse. Other implementation details such as hyper-parameters and the training scheme can be found in Appendix~\ref{sec:appendix_setup}.}
\end{remark}

\begin{remark}[\QGG.] This model learns to generate query graphs from question-answer pairs using reinforcement learning. BERT is also used but only to get a matching score between question and logical form via joint encoding, which is served as one of the seven hand-crafted features used for ranking. To train this model on \OurDataset, we use the same configuration from the original paper for \WebQSP, i.e., considering up to 2-hop relations in the KB for candidate generation. This can cover 80\% of the training questions in \OurDataset. Increasing it to 3 hops will take a few months (estimated) to train this model due to the large number of entities in \OurDataset. We only report F1 for \QGG because it uses a different meaning representation.
\end{remark}

\subsection{Results}

\subsubsection{Overall Evaluation}
%We present our experiment results here together with some insights drawn from the results. As depicted in table~\ref{table:cross}, for the cross-domain setting, the performances of two generation models, i.e., Seq2Seq and Grammar model, are both very low. Seq2Seq can only achieve an exact match of 5.2\%, while Grammar model is a little bit higher than Seq2Seq as it imposes constraints to guarantee well-formed logical forms during decoding. By comparison, Seq2Seq \Ranking outperforms Seq2Seq significantly, which reveals the importance of taking account of the underlying structure of KB as prior knowledge. Note that, when incorporating BERT, the performance of Seq2Seq improves from 5.2\% to 37.1\%, and the performance of Seq2Seq \Ranking improves from 48\% to 65.4\%. We argue that the reason for this is that BERT encodes much knowledge from the external corpus that can be transferable to any domain.

We show the overall results in Table~\ref{table:overall}. \Ranking achieves the best overall performance on \OurDataset. Both of our models outperform \QGG, demonstrating their competitive performance. We also observe a significant performance drop on all the variants of our models, which suggest that both BERT encoding and VP play an important role.

Next we drill down to different levels of generalization. The results clearly demonstrate the key role of contextualized encoding via BERT for compositional and zero-shot generalization: While BERT and GloVe perform similarly in \iid generalization, using GloVe instead of BERT, \Transduction's F1 drops by 17.5\% in compositional generalization and by 26.2\% in zero-shot generalization. For \Ranking, it also brings a 21.9\% drop in zero-shot generalization. We do a more in-depth analysis on \Transduction and confirm that this is mainly because BERT enables better generalization to unseen schema items. For example, for question \nl{What home games did the Oakland Raiders play?} about an unseen domain \wl{american\_football}, \Transduction can find the correct relation \wl{american\_football.football\_game.home\_team}, while \Transduction without BERT is confused by the wrong relation \wl{sports.sports\_team.arena\_stadium}, which is seen during training. This is a common type of error by GloVe-based models but much less frequent by BERT-based models, which shows how pre-trained contextual embeddings help address the \emph{language-ontology alignment} challenge. Since \QGG is also a ranking model that uses BERT, it is not surprising it also performs reasonably in compositional and zero-shot generalization.

On the other hand, \Ranking outperforms \Transduction significantly in compositional and zero-shot generalization. This is largely due to \Ranking's effectiveness in \textit{search space pruning}, i.e., \Ranking prunes the search space based on the each identified entity's neighboring facts, while \Transduction only uses the ontology, which is less discriminating. VP helps \Transduction to some extent, but still not quite up to the same level of effective pruning \Ranking enjoys. This also provides a plausible explanation for the interesting observation that BERT seems to play a less significant role in compositional generalization in ranking models than in transductive models --- because of the effective search space pruning, \Ranking's candidate set includes much less of the logical forms seen in training that may be confusing to the model. 

\nop{
Finally, we also observe that all settings of \Transduction perform well on \iid but Transudction without VP, this is because when training BERT-based model, we can only afford to train the model with a subset of the vocabulary using VP, not using VP during inference will lead to a mismatch. 
}

\nop{
Finally, we notice that the \QGG~\cite{lan-jiang-2020-query} performs uniformly across 3 levels of generalization. A possible explanation is that \QGG is trained from question answer pairs instead of question logical form pairs. There can be many different queries lead to the same answer, which might enable \QGG to explore more different patterns during training.
}

%1. VP is more useful for cross-domain setting. This is because it is more challenging to identify the relevant ontology items for cross-domain setting.

%2. Seq2Seq use the entire vocabulary for training while Seq2Seq + BERT only use a subset of vocabulary to train. This explains the performance on realistic setting, and also indicates the optimal way for using BERT in KBQA should be further studied.

%3. An interesting phenomenon we observed is that BERT can significantly improve the performance for cross-domain setting, but not the realistic setting.

%4. \Ranking model performs much better than generation model, which reveals the importance of taking account of the underlying structure of KB.
 
\begin{figure}[!ht]
\centering
\includegraphics[width=0.9\linewidth]{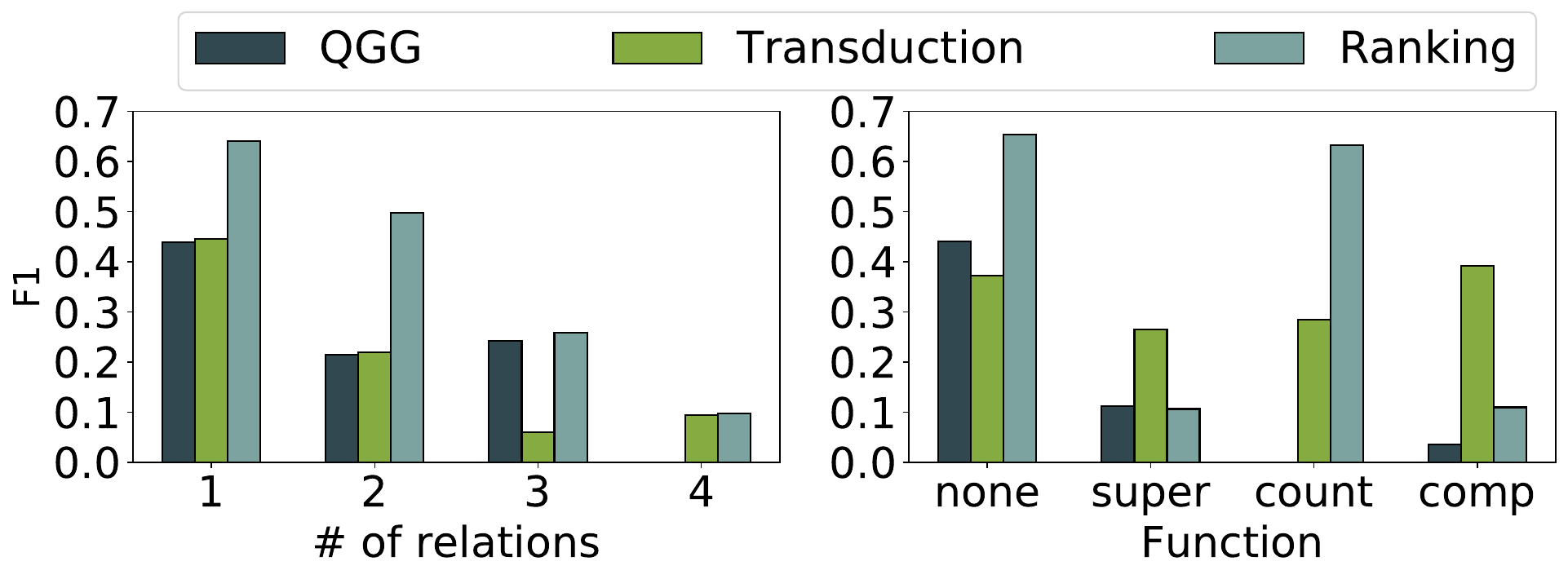}
\caption{Fine-grained results.}
\label{fig:fine}
\vspace{-10pt}
\end{figure}

\subsubsection{Fine-Grained Evaluation}
\label{sec:fine-grained}
We now present a more fine-grained analysis along several other dimensions in~\autoref{fig:fine}.

\begin{remark}[Structural complexity.] The performance of all models degrades rapidly as questions become more complex. Note that, though our \Ranking model only generates candidates with up to 2 relations, it still has a chance to get partially correct on more complex questions. In fact, it even outperforms the more flexible \Transduction model on questions with 3 relations, which again demonstrates the importance of effective search space pruning.
\end{remark}

\begin{remark}[Function.] Ranking models (including \Ranking and \QGG) rely on topic entities for candidate generation which are often absent in questions with comparatives or superlatives, e.g., \nl{which chemical element was first discovered?} Their performance suffers as a result. \QGG gets zero F1 on questions that requires counting since it only returns entities as answer. On the other hand, \Transduction performs significantly better than the other two models on superlatives and comparatives due to its flexibility in generating all types of logical forms, though that comes with a penalty on simpler questions. Better guided search for both \Ranking and \Transduction may be a promising future direction to enjoy the best of both worlds.
\end{remark}

\begin{figure}[t]
\centering
\includegraphics[width=0.9\linewidth]{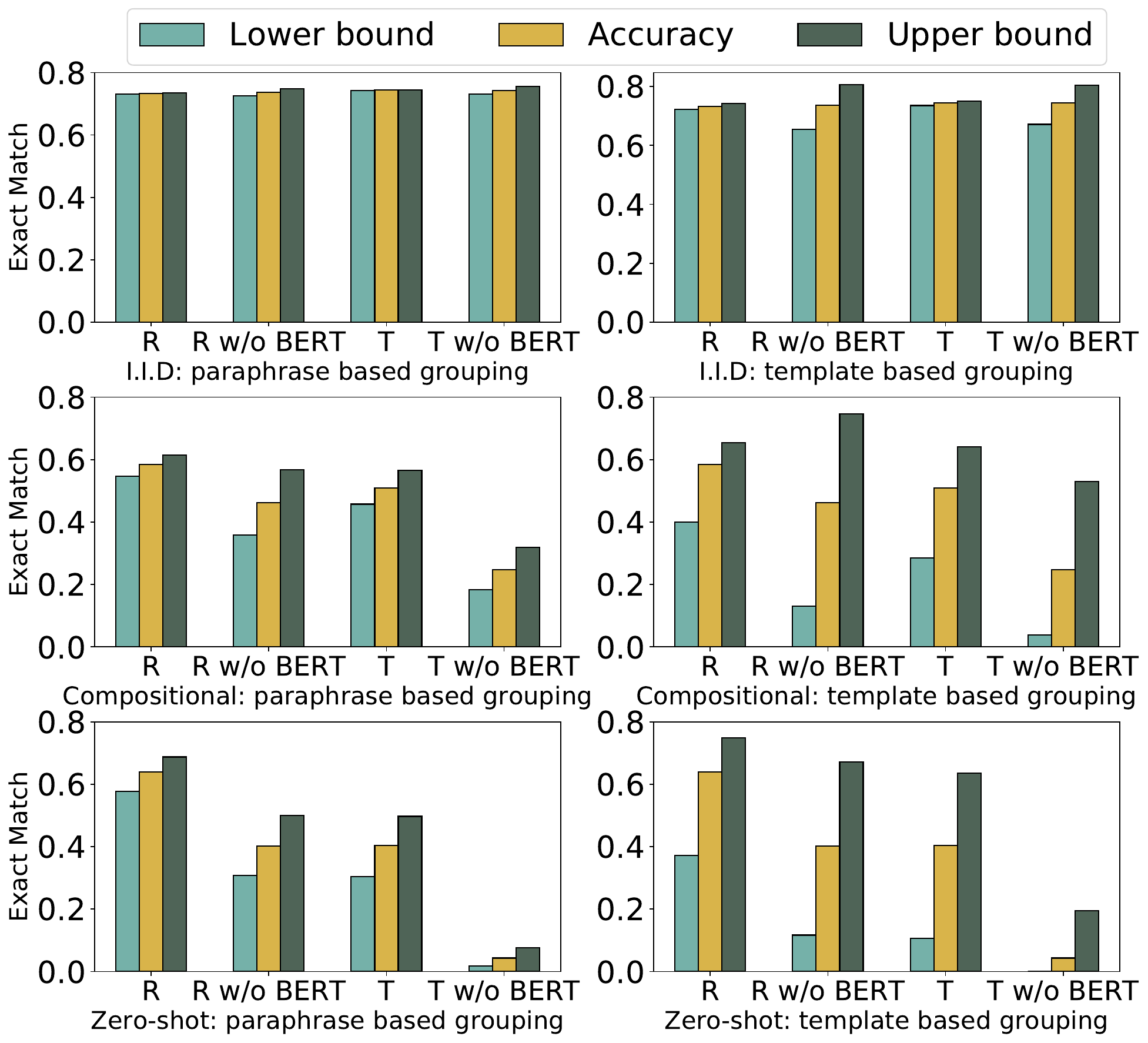}
\caption{Robustness analysis. T denotes \Transduction and R denotes \Ranking. For robustness analysis, we assume correct entities are given.}
\label{fig:para}
\vspace{-10pt}
\end{figure}

\subsubsection{Robustness Analysis} A good \Problem model should be robust to different paraphrases and entity groundings (e.g., should not succeed at \nl{Where is the Trump tower?} but fail at \nl{Where is the Tune Hotels?}). \OurDataset provides an opportunity to directly test that. We create different groupings of questions: \emph{grouping by paraphrase} --- questions that are the same except for their entity groundings fall into the same group, and \emph{grouping by template} --- questions from the same canonical logical form but with different groundings or paraphrases are grouped together. Each grouping provides upper and lower bounds for evaluating model robustness. The upper bound is derived by treating all the questions in a group as correct if any of them is correctly answered, while the lower bound is derived by treating all the questions as incorrect if any of them is incorrectly answered. The true accuracy should lie between the bounds. The closer the three are, the more robust a model is to the corresponding mutation. To reduce the number of variables in this analysis, we assume perfect entity linking and eliminate entity linking errors. The results are shown in Figure~\ref{fig:para}, which clearly show that BERT-based models are generally more robust than GloVe-based models (reflected by the smaller gap between their upper and lower bounds) and provides further explanation to its superior performance in non-\iid generalization. It is interesting to observe that the \Ranking model yields a higher upper bound with GloVe than with BERT in non-\iid generalization, which shows that the major problem with GloVe is its robustness to variations.

\nop{A concrete example is as follows, \Ranking without BERT can get correct on \nl{Which state's capital is raipur?} but fails on \nl{Which state's capital is bhubaneswar?}. On the other hand, \Ranking can get both of them correct.}

\subsection{Error Analysis}
We analyze 100 randomly sampled questions that our \Ranking model gets wrong to discuss venues for future improvement. We summarize the errors into the following categories.

\begin{remark}[Coverage limitation (34\%):] Due to the diversity of questions in \OurDataset in terms of both complexity and functions, candidates generated by \Ranking can only cover around 80\% of the questions, as it can only enumerate up to 2-relational logical forms due to combinatorial explosion. Uncovered questions constitute 34\% of errors. More intelligent ways for search and candidate generation are promising future directions.
\end{remark}

\begin{remark}[Entity linking (33\%):] The large-scale mining of entity surfaces forms makes entity linking a unique challenge in \OurDataset. For example, in \nl{After apr. the 2nd, 2009, which song was added to RB?}, \nl{RB} refers to the music video game Rock Band, while our entity linker fails to identify any entity from this question. Also, less popular entities form another source of entity linking errors. To be specific, popularity is a very strong signal for entity disambiguation in previous datasets, but not as strong in \OurDataset because many questions are intentionally not just about very popular entities. For \nl{What is the belief of Mo?} our entity linker can identify the entity mention \textit{Mo}, and the most popular entity that has surface form \textit{Mo} is the state of Missouri, while the question is asking about the Mo religion. Accurate entity linking on large-scale KBs is still a major challenge.
\end{remark}

\begin{remark}[Relation mis-classification (26\%):] Another major type of errors comes from misclassified relations. Precisely understanding a vaguely or implicitly expressed relation is challenging, especially for zero-shot generalization. For example, question \nl{Which video game engine is the previous version of Frostbite 2?} is from an unseen domain \wl{cvg}, and the correct relation for this question should be \wl{cvg.computer\_game\_engine.successor\_engine}, while \Ranking predicts \wl{cvg.computer\_game\_engine.predecessor\_engine}, which fails to capture the alignment between \wl{successor\_engine} and \nl{previous version}. Besides, around 20\% of relation mis-classifications are due to that the model cannot correctly determine the answer type. For question \nl{Name the soundtrack that Amy Winehouse features.}, which is asking about a music soundtrack, but \Ranking predicts a relation associated with the answer type \wl{music.album}.
\end{remark}

\begin{remark}[Other (7\%):] The rest of the errors mainly include typing errors and function errors. For typing errors, the question can seek for information about more specific types (e.g., looking for \wl{politician} instead of just \wl{person}), while \Ranking does not explicitly handle type constraint as introducing type constraints by enumeration will result in prohibitively many candidate logical forms. A more flexible way of handling type constraint is in need. Also, sometimes \Ranking can be confused by the function of a question. For instance, for question \nl{Name the image which appears in the topic gallery armenian rock.}, there shouldn't be any aggregation function, however, \Ranking mistakenly predicts a counting function for it.
\end{remark}

\subsection{Transfer Learning}
\label{sec:transfer}

We also show that \OurDataset could serve as a valuable pre-training corpus for \Problem in general by pre-training \Ranking on \OurDataset and testing its transferability to \WebQSP, which contains naturally-occurring questions from Google search log. To adapt \Ranking to \WebQSP,  we first convert SPARQL queries in \WebQSP to our S-expression logical forms. We test three settings: Full training that fine-tunes using all training data of \WebQSP, few-shot that only uses 10\%, and zero-shot that directly applies models trained on \OurDataset to \WebQSP. For simplicity, we assume perfect entity linking. The results are shown below. Pre-trianing on \OurDataset uniformly improves the performance, especially in the low-data regime. Most remarkably, pre-trained on \OurDataset, \Ranking can achieve an F1 of 43\% without \textit{any} in-domain training on \WebQSP. These results provide further supporting evidence for the quality and diversity of \OurDataset.

\begin{table}[!h]
    \small
    \centering
    \resizebox{0.3\textwidth}{!}{
    \begin{tabular}{c|ccc}
    \toprule
     \multicolumn{1}{c}{}&\multicolumn{1}{c}{}&\multicolumn{1}{c}{\textbf{Exact Match}}& \textbf{F1} \\ \midrule
    %  \textbf{Full Training} \\ \midrule
       \multirow{2}{*}{\textbf{Full Training}}
      &\textbf{\Ranking}     & 0.59 & 0.67\\
      &\multicolumn{1}{r}{+pre-training}   & \textbf{0.60} & \textbf{0.70} \\\midrule
    %   \textbf{Few-shot} \\ \midrule
     \multirow{2}{*}{\textbf{Few-shot}}
      &\textbf{\Ranking}  & 0.44 & 0.53\\
      &\multicolumn{1}{r}{+pre-training} &\textbf{0.55} & \textbf{0.65}\\\midrule
    %   \textbf{Zero-shot}\\ \midrule
    \multirow{2}{*}{\textbf{Zero-shot}}
      &\textbf{\Ranking}  & 0.00 & 0.00\\
      &\multicolumn{1}{r}{+pre-training} & \textbf{0.35} & \textbf{0.43}\\
      
    %  \multirow{2}{*}{\textbf{Zero-shot}}& \textbf{s2s+BERT+VP} & 0.14 & 0.20 \\
    %   &\textbf{\Ranking+BERT} & \textbf{0.35} & \textbf{0.43} \\
    \bottomrule
    \end{tabular}}
    \nop{\caption{Transfer Learning Results on \WebQSP.\nop{ "Pre-training" means pre-training on \OurDataset, and the base method is directly fine-tuning from BERT.}}
    \label{table:transfer}}
    \vspace{-10pt}
\end{table}

\section{Related Work}
%Briefly compare GraphQuestions 2.0 with some existing KBQA datasets. We need to make it clear that current datasets are not good enough in terms of complexity, quality and quantity.\\
%We may also need to address the paper by Jonathan Berant here (i.e., don't paraphrase! detect).
%Executable semantic parsing~\cite{executable} converts natural language utterances into queries that can be directly executed to get the denotations. It has attracted most of the research interest in the field of semantic parsing. In the following of this paper, we will use executable semantic parsing and semantic parsing interchangeably.\\

\begin{remark}[Existing \Problem datasets.]
There have been an array of datasets for \Problem in recent years. \WebQ~\cite{berant-etal-2013-semantic} is collected from Google search logs with the answers annotated via crowdsoursing. \citet{yih-etal-2016-value} later provide logical form annotation for the questions in \WebQ. Most of the questions in \WebQ are simple single-relational questions, and \citet{talmor-berant-2018-web} generate more complex questions by automatically extending the questions in \WebQ with additional SPARQL statements, leading to the \ComplexQ dataset. However, such automatic extension may lead to unnatural questions. \SimpleQ~\cite{DBLP:journals/corr/BordesUCW15} is another popular \Problem dataset created by sampling individual facts from \Freebase and annotating them as natural language questions. It therefore only contains single-relational questions. \GraphQ~\cite{su-etal-2016-generating} is the most related to ours. It proposes an algorithm to automatically generate logical forms from large-scale KBs like \Freebase with guarantees on well-formedness and non-redundancy. We also use their algorithm for logical form generation, and develop a crowdsourcing framework which significantly improves the scalability and diversity. The aforementioned datasets are all based on \Freebase. \QALD~\cite{qald9} and \QuAD~\cite{trivedi2017lc} are popular datasets on \DBpedia. \QALD is manually created by expert annotators, while \QuAD first generates SPARQL queries and unnatural canonical questions with templates and have expert annotators paraphrase the canonical questions.

Most \Problem datasets primarily operate with the \iid assumption because their test set is a randomly sampled subset of all the data~\cite{berant-etal-2013-semantic,yih-etal-2016-value,DBLP:journals/corr/BordesUCW15,talmor-berant-2018-web,trivedi2017lc}. \GraphQ~\cite{su-etal-2016-generating} is set up to primarily test compositional generalization by splitting their training and test sets on logical forms, similarly for \QALD~\cite{qald9}. However, none of the existing datasets supports the evaluation of all three levels of generalization. Our dataset also compares favorably to existing datasets in other dimensions such as size, coverage, diversity (\autoref{table:dataset}).
\end{remark}

\nop{
\WebQ~\cite{berant-etal-2013-semantic,yih-etal-2016-value} is collected from Google query logs, which mainly comprises simple questions. The entire dataset is randomly split into training and test set with no other control, as a result, vast majority of the logical forms in test data are also seen in training. It means \WebQ only tests \iid generalizability of \Problem models. \ComplexQ~\cite{DBLP:journals/corr/abs-1803-06643} extends simple questions in \WebQ to complex questions following several rules (e.g., adding more constraints and adding superlatives). Ideally, \ComplexQ should be able to test compositional generalizability of \Problem models, however, due to its random split, the canonical logical forms of around 72\% questions in test set are seen in training, which makes \ComplexQ mostly an \iid \Problem dataset. Questions are drawn from different distributions for training and test in \QALD~\cite{qald9} and \GraphQ~\cite{su-etal-2016-generating}. These two datasets can be used to test the second level of generalizability, however, for \QALD, most relations in test set are also seen in training, so it fails to meet the requirement for testing zero-shot generalizability. \GraphQ also doesn't explicitly constrain questions in test set to be from new domains, which makes it less ideal to test zero-shot generalizability. Moreover, the limited scale of \QALD and \GraphQ further undermines their roles in benchmarking modern neural models. A more recent dataset is \CFQ~\cite{keysers2019measuring}, which is a synthetic dataset focusing on compositional generalizability. It cannot be used to test zero-shot generalizability either. As a result, it is indispensable to have a new large-scale \Problem dataset of high quality that can test all 3 levels of generalizability.
}

\begin{remark}[Existing models on \Problem.] 
\Problem models can be roughly categorized into semantic-parsing based methods and information-retrieval methods. The former maps a natural language utterance into a logical form that can be executed to get the final answer, while the latter directly ranks a list of candidate entities without explicitly generating a logical form. We focus on semantic-parsing methods due to their superior performance and better interpretability. Existing methods are mainly focused on \iid setting and are limited in generalizability. Conventional rule-based methods~\cite{berant-etal-2013-semantic} suffer from the coverage of their hand-crafted rules and therefore have rather limited generalizability. More recent models either employ encoder-decoder framework~\cite{jia-liang-2016-data, dong-lapata-2016-language, zhang-etal-2019-complex} to decode the logical form auto-regressively or first generate a set of candidate logical forms according to predefined templates or rules, and then match each candiate with the utterance to get the best matched one\cite{berant-etal-2013-semantic,yih-etal-2015-semantic, kbqa, reddy-etal-2017-universal, Abujabal, luo-etal-2018-knowledge,maheshwari2019learning, sun2020sparqa, lan-jiang-2020-query, diefenbach2020towards, hua2020less}. These models already achieve impressive results on \iid dataset like \WebQSP, however, on \GraphQ that mainly tests compositional generalization, the best model can only achieve an F1 of 21.5~\cite{sun2020sparqa}. This demonstrates that non-\iid generalization in \Problem has not drawn enough attention from existing methods.
\end{remark}

% \noindent \textbf{Semantic parsing.}\nop{ We have discussed efforts on both data and modeling for OpenSP. Here we situate this work in the broader literature of semantic parsing.} Semantic parsing started off as natural language interfaces to databases~\cite{woods1973progress,geo1996,zettlemoyer2005learning}, which recently has received renewed interests rebranded as text-to-SQL parsing~\cite{wikisql,yu-etal-2018-spider}. Each question is provided with a small table or database (e.g., dozens of schema items and hundreds of records). Initiatives such as semantic web and Internet of Things, and the increasing heterogeneity and interconnectedness of computing in general, have motivated the development of OpenSP~\cite{berant-etal-2013-semantic,kwiatkowski-etal-2013-scaling,cai-yates-2013-large}. We re-examine the OpenSP problem through the lens of its unique challenges, and correspondingly provide data, models, and analyses to facilitate further development. Also related is broad-coverage semantic parsing such as AMR~\cite{banarescu-etal-2013-abstract}, with the difference being OpenSP's semantics is grounded, e.g., to a knowledge base. 

\begin{remark}[Pre-training and non-\iid generalization.]
The problem of non-\iid generalization has drawn attention under related semantic parsing settings. \citet{su-yan-2017-cross} recognize the key role of pre-trained word embeddings~\cite{mikolov2013distributed} in cross-domain semantic parsing. Contextual embeddings like BERT are later shown to be successful for cross-domain text-to-SQL parsing~\cite{hwang2019comprehensive}. Another line of work has been focusing on compositional generalization on a number of specifically-synthesized datasets~\cite{lake2018generalization,keysers2019measuring}. Concurrent to this work, \citet{furrer2020compositional} find that pre-trained contextual embeddings plays a more vital role in compositional generalization than specialized architectures. To the best of our knowledge, our work is among the first to demonstrate the key role of contextual embeddings like BERT at multiple levels of generalization for \Problem. 
% We also show that our dataset can be used as pre-training to enable few-shot and zero-shot generalization to other \Problem datasets, which is in line with recent development on pre-trained models as few-shot learners~\cite{brown2020language}.
\end{remark}

\section{Concluding Remarks}
\label{sec:conclusion}
In this paper, we explicitly lay out and study three levels of generalization for \Problem, i.e., \iid, compositional, and zero-shot generalization. We construct and release \OurDataset, a large-scale, high-quality \Problem dataset with 64,331 questions that can be used to evaluate all three levels of generalization. We also propose a novel BERT-based \Problem model. The combination of our dataset and model enables us to thoroughly examine and demonstrate, for the first time, the different challenges and promising solutions in non-\iid generalization of \Problem. In particular, we demonstrate the key role of pre-trained contextual embeddings like BERT in compositional and zero-shot generalization of \Problem.

This work is just a starting point towards building more practical \Problem models with stronger generalization. It opens up a number of future directions. First, for a full-fledged QA system on large-scale KBs, more sophisticated, context-sensitive entity linkers that can work for long-tail entities and variation in surface forms are needed. Entity linking is of particular importance because topic entities provide important anchors in the large KB to dramatically prune the search space, but is also more challenging due to the large number of entities. Second, for complex questions, the search space is still prohibitively large if we enumerate candidate logical forms in a brute-force way. More intelligently guided search is a promising future direction to efficiently generate the most promising candidates and prune less promising ones. Last but not least, even though we have empirically demonstrated how pre-trained contextual embeddings like BERT significantly facilitate compositional and zero-shot generalization by providing better language-ontology alignment, there is still a lack of deeper understanding on why that is the case, which may inspire better ways of exploiting these models. We are also interested in experimenting with other pre-trained contextual embeddings such as RoBERTa~\cite{liu2019roberta} and BART~\cite{lewis2019bart}. 

\nop{
With \OurDataset being a testbed, we further propose a line of baseline models that enable us to conduct comprehensive evaluations. Based on the results from our baseline models on \OurDataset, we identify \textit{pre-trained contextual embeddings} and \textit{effective search space pruning} are two key factors that would lead to better generalizability. Still, the performance of all the baseline models, including a state-of-the-art model on existing datasets, are far from ideal, which suggest a great venue for future improvement. Future directions for improvement suggested in our experiments include: better entity linking system that can work for more realistic surface form and less popular entities, more flexible way of pruning search space and better language understanding under zero-shot setting. We hope \OurDataset can be a stepping stone towards more advanced \Problem models with stronger performance and better generalizability in the future.
}
% In this paper, we study the problem of non-\iid generalization in \Problem, which is an important problem that has been under-addressed by most existing works. Specifically, we introduce three different levels of generalizability of practical \Problem models, namely, \iid, compositional and zero-shot, to qualitatively study this problem. 
% Based on our new large-scale dataset \OurDataset and comprehensive evaluations with our proposed models and also a state-of-the-art model on existing datasets, we are able to identify \textit{pre-trained contextual embeddings} and \textit{effective search space pruning} are two key factors that would lead to successful non-\iid generalization. 

\section{Acknowledgements}
We would like to thank Jefferson Hoye for helping with the crowdsourcing pipeline, the graduate students who helped with canonical question annotation, and the anonymous reviewers for their helpful comments.
This research was sponsored in part by NSF 1528175, a Fujitsu gift grant and Ohio Supercomputer Center \cite{OhioSupercomputerCenter1987}. 
% The views and conclusions contained herein are those of the authors and should not be interpreted as representing the official policies, either expressed or implied, of the Army Research Office or the U.S. Government. The U.S. Government is authorized to reproduce and distribute reprints for Government purposes notwithstanding any copyright notice herein.}

%%
%% The acknowledgments section is defined using the "acks" environment
%% (and NOT an unnumbered section). This ensures the proper
%% identification of the section in the article metadata, and the
%% consistent spelling of the heading.
% \begin{acks}
% To Robert, for the bagels and explaining CMYK and color spaces.
% \end{acks}

%%
%% The next two lines define the bibliography style to be used, and
%% the bibliography file.
\bibliographystyle{ACM-Reference-Format}
\bibliography{www2021}

%%
%% If your work has an appendix, this is the place to put it.
\appendix

\section{Crowd Worker Demographics}
\label{sec:appendix_demographics}

\begin{figure}[!h]
    \centering
    \begin{subfigure}{0.45\columnwidth}
        \centering
        \includegraphics[width=\linewidth]{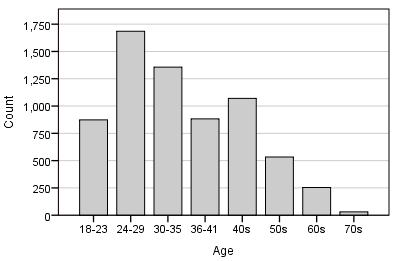}
        % \caption{1a}
        % \label{fig:sfig1}
    \end{subfigure}%
    \begin{subfigure}{0.45\columnwidth}
        \centering
        \includegraphics[width=\linewidth]{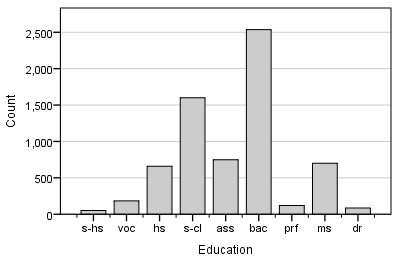}
        % \caption{1a}
        % \label{fig:sfig1}
    \end{subfigure}%
    \vspace{-5pt}
    \caption{
    Demographics of participating crowd workers: age (top) and level of education (bottom). s-hs = some high school, voc = vocational school, hs = high school, s-cl = some college, ass = associate degree, bac = bachelor degree, prf = professional degree, ms = master degree, dr = doctoral degree.
    }
    \label{fig:demographics}
    \vspace{-10pt}
\end{figure}

\noindent Our data collection included 6,685 crowd workers recruited through Amazon Mechanical Turk (AMT). Each worker completed a short anonymous demographic survey requesting information on their gender, age, and completed level of education. Gender representation was fairly even, with only slightly higher participation by females (59.1\%) than by males (40.9\%). The crowd workers also exhibited a diverse demographic in terms of both age and level of education, as shown in~\autoref{fig:demographics}. These facts provide a strong indication of the high degree of diversity in this dataset.
\vspace{-0.1in}
\begin{figure}[t]
\centering
\includegraphics[width=1.05\linewidth]{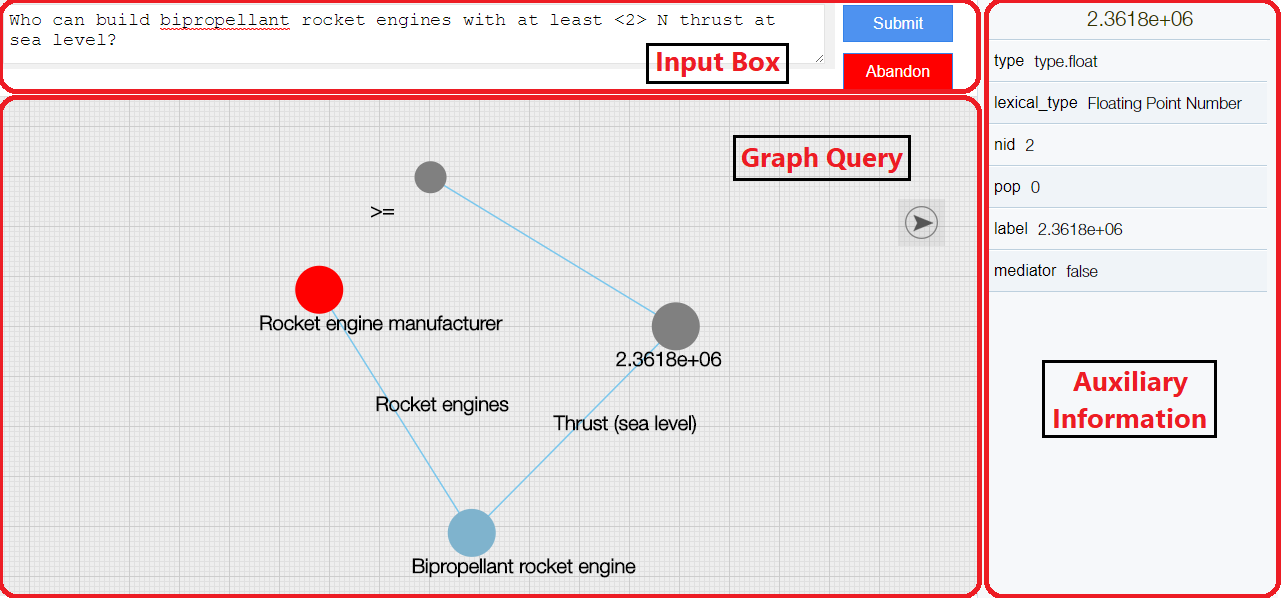}
\caption{Interface for canonical question annotation.}
\label{fig:annotation}
\vspace{-10pt}
\end{figure}

\nop{
The study included 6,685 crowd workers recruited through Amazon Mechanical Turk (AMT).  All crowd workers reported being U.S. citizens and 18 years of age or older.  Only crowd workers with a minimum 95\% task approval rating were provided access to the AMT Human Intelligence Task (HIT) for all three tasks (paraphrasing, cross-validation, and entity mention selection).  These tasks required reading English; therefore, proficiency in English was a requirement stated in the HIT description and 98.1\% of the crowd workers reported English as their native language.  Each worker completed a short demographic survey requesting information on their gender, age, and completed level of education. 

Gender representation was fairly even, with only slightly higher participation by females (59.1\%) than by males (40.9\%); and, crowd workers exhibited a diverse demographic in terms of both age and level of education as well.  The survey listed eight age categories (18-23, 24-29, 30-35, 36-41, 40s, 50s, 60s, and 70s), all of which were represented in the survey responses, with the highest percentage (20.3\%) falling in the 30-35 years of age category.  Figure~\ref{fig:demographics} (top) shows the number of responses for each age category.  

The survey listed 11 levels of education (none, other, some high school, vocational school, high school, some college, associate degree, bachelor degree, professional degree, master degree, and doctoral degree).  Similar to age, all education levels were represented in the survey responses.  Note that the education level categories of ‘none’ and ‘other’ were reported by only four crowd workers.  These categories were, therefore, conflated with the category for ‘vocational school’, which reduced the number of level-of-education categories to nine.  The highest percentage of workers reported completion of a Bachelor’s degree (37.9\%).  Figure~\ref{fig:demographics} (bottom) shows the number of responses for each level-of-education category.

The facts, not only that the workers on this project represent all categories for both age and level-of-education demographics, but also that there is generally equivalent participation by gender is a strong indication of the high degree of diversity in this dataset.
}

\nop{
\subsection{Interface for Canonical Question Annotation}
\label{sec:appendix_annotation}

\autoref{fig:annotation} shows the interface used by graduate students to validate logical forms and annotate canonical questions. It contains a graphical visualization of the logical form, where one can click on each node and edge to have auxiliary information such as type, and node id, and description displayed on the right. If a logical form is deemed as not meaningful, i.e., we could not reasonably expect it to be asked by real users, it is abandoned by clicking the Abandon button; otherwise the annotator writes a corresponding natural language question with placeholders for entities and literals. }

\begin{figure}[htbp]
    \centering
    \begin{subfigure}{\linewidth}
        \centering
        \includegraphics[width=\linewidth]{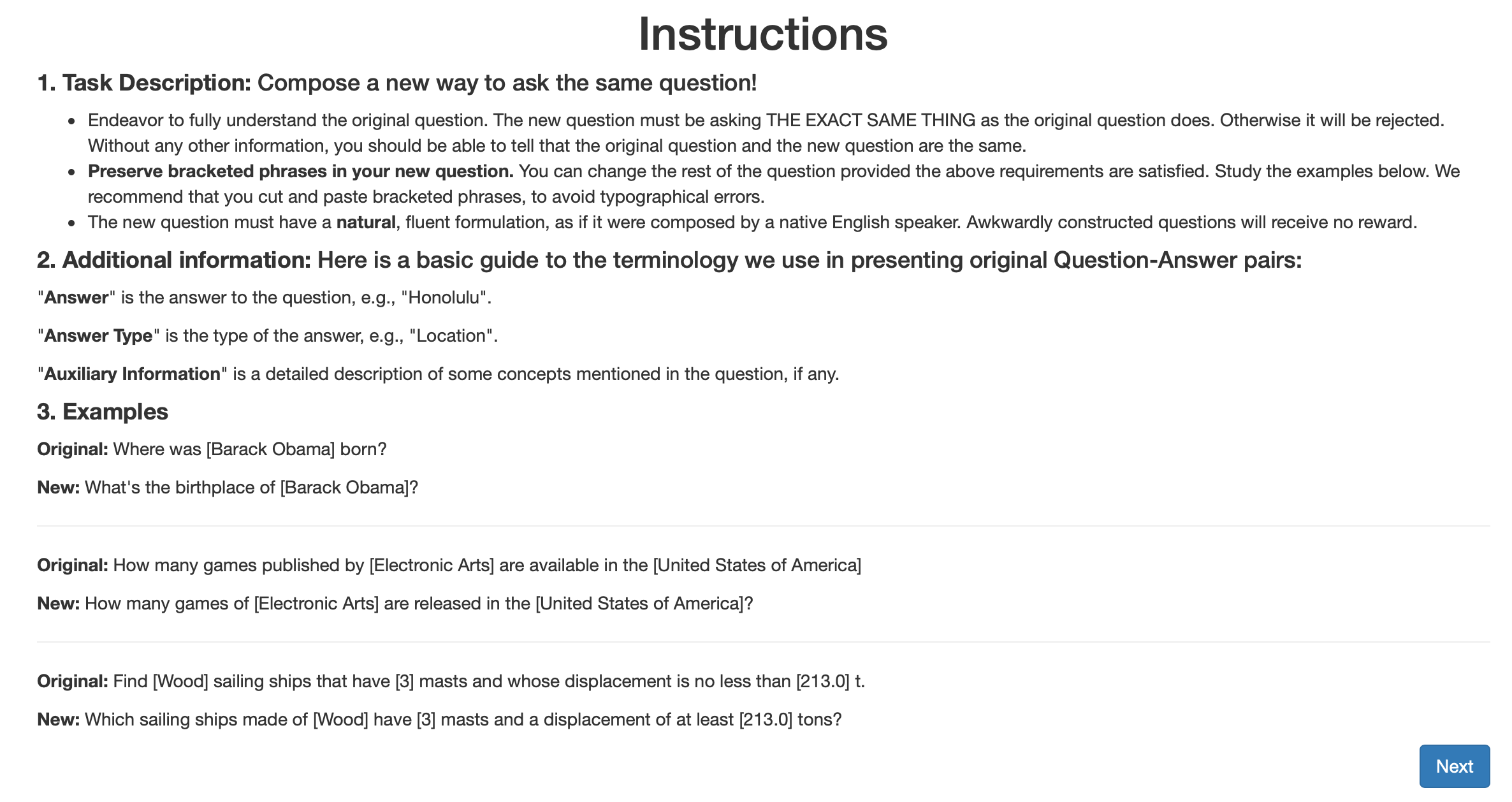}
        \caption{Task 1: Paraphrasing.}
        % \label{fig:sfig1}
    \end{subfigure} \\
    \begin{subfigure}{\linewidth}
        \centering
        \includegraphics[width=\linewidth]{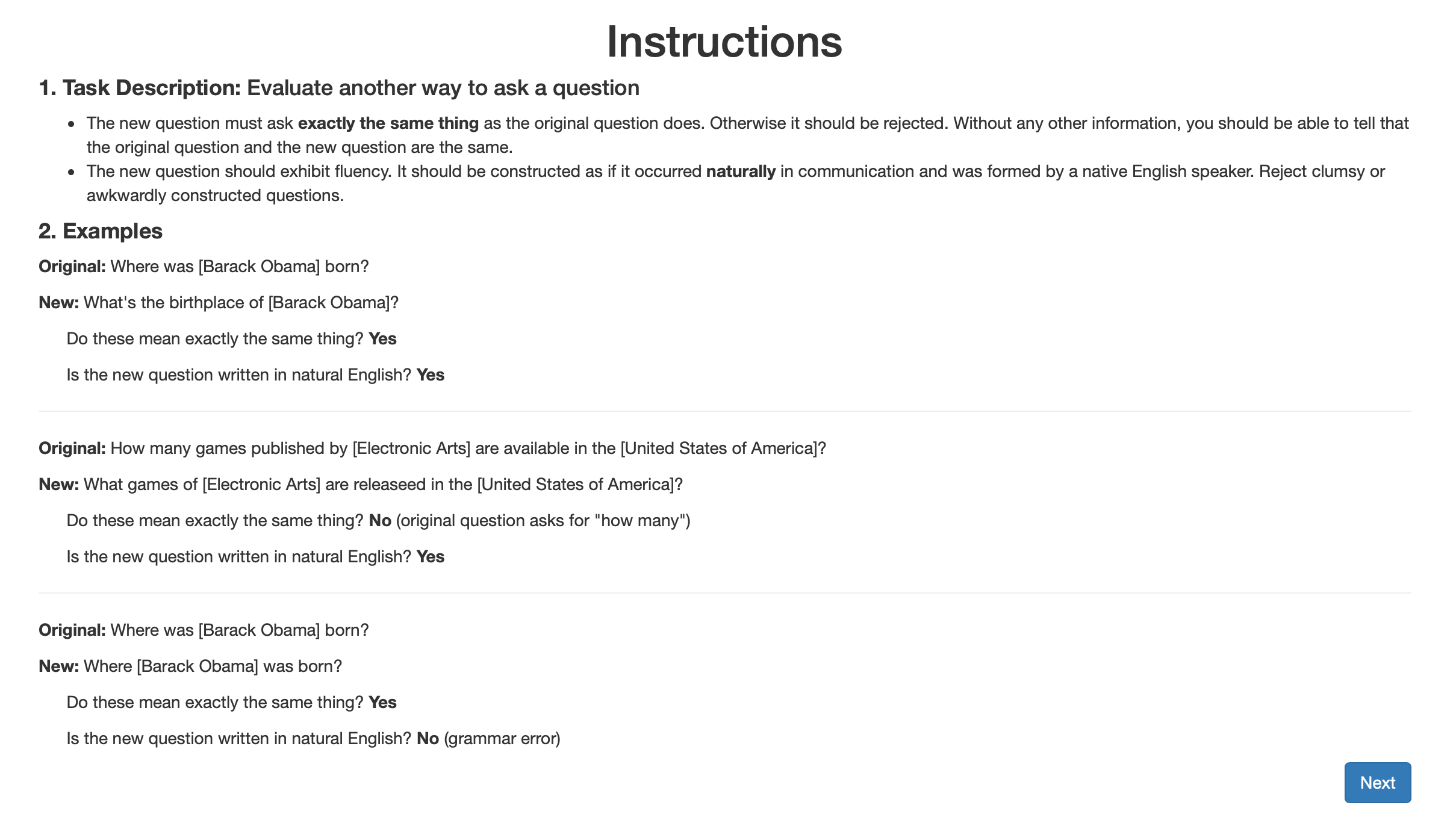}
        \caption{Task 2: Cross-validation.}
        % \label{fig:sfig1}
    \end{subfigure} \\
    \begin{subfigure}{\linewidth}
        \centering
        \includegraphics[width=\linewidth]{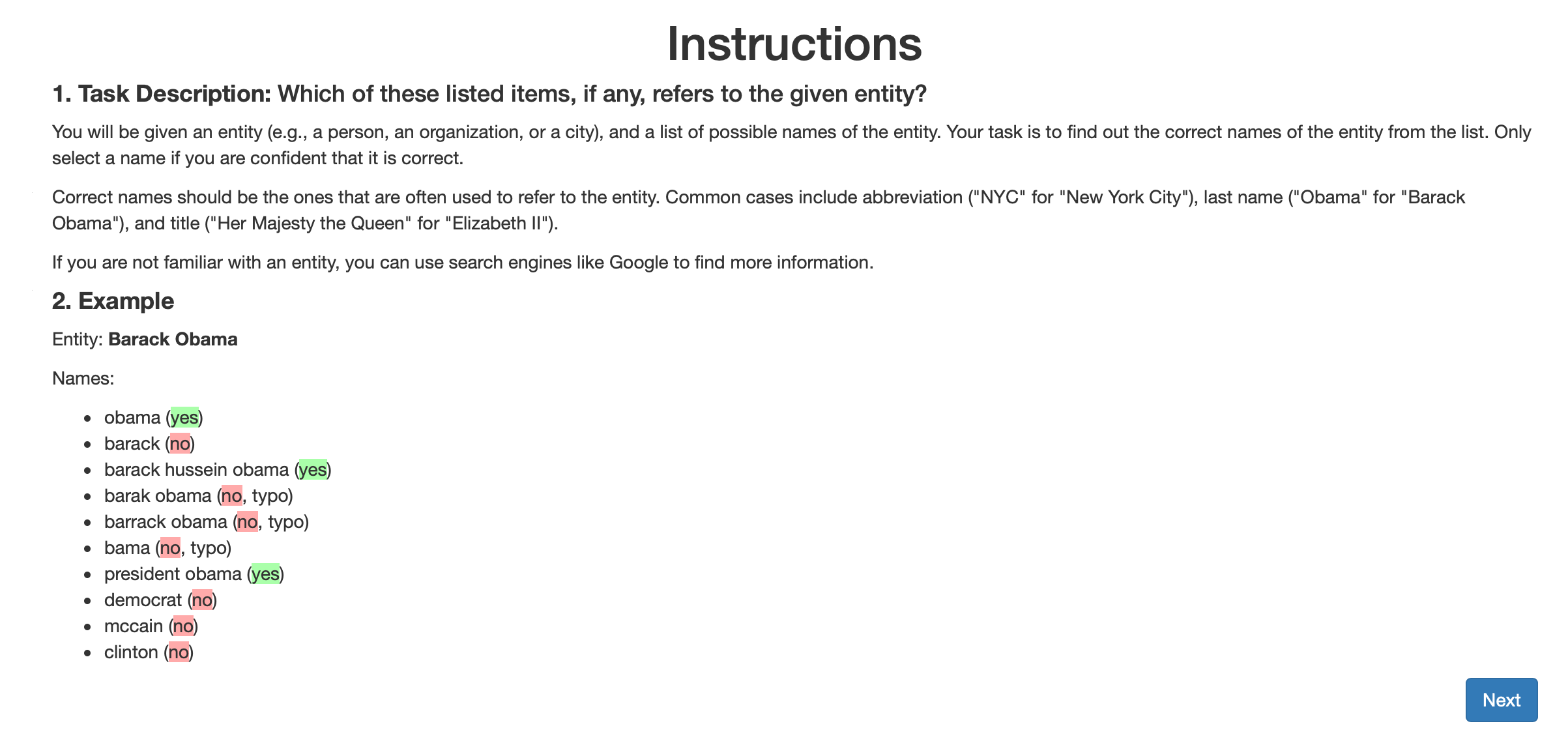}
        \caption{Task 3: Entity surface form mining.}
        % \label{fig:sfig1}
    \end{subfigure}%
    \caption{
    Crowdsourcing instructions.
    }
    \label{fig:crowdsourcing_instructions}
\end{figure}

\nop{
\subsection{Instructions for Crowdsourcing}
\label{sec:crowdsourcing_instructions}

The instructions for each of the crowdsourcing tasks are shown in \autoref{fig:crowdsourcing_instructions}.
}

% \subsection{Distribution of Questions}\label{sec:appendix_dist}
% To show the diversity of questions in our dataset, we make a sunburst chart to visualize the distribution of first 3 words in questions, only cases that shared by over 100 questions are labeled by us. As shown in Figure~\ref{fig:dist}, questions in our dataset is very diverse thanks to the effort made by crowd workers. \textit{What} questions are most frequently asked in our data, which is expected for dataset on factoid question answering. 

% \subsection{Linguistic Diversity of Questions}\label{sec:appendix_dist}
% To show the diversity of questions in our dataset, we make a sunburst chart to visualize the distribution of first 3 words in questions, only cases that shared by over 100 questions are labeled by us. As shown in Figure~\ref{fig:dist}, questions in our dataset is very diverse thanks to the effort made by crowd workers. \textit{What} questions are most frequently asked in our data, which is expected for dataset on factoid question answering. 

\section{S-expression}\label{sec:lisp}
Our S-expressions employ set-based semantics: Each function takes a number of arguments, and both the arguments and the denotation of the functions are either a set of entities or entity tuples. Function definitions are listed in Table~\ref{table:lisp}. There are 3 argument types in total, namely, a set of entities, a set of (entity, entity) tuples and a set of (entity, value) tuples. Classes and single entities are the most fundamental examples of set of entities, and relations are the most fundamental examples of set of binary tuples. By applying those functions defined in our grammar, we are able to get more complex set of entities and binary tuples.
\intextsep 0.05in
\begin{table}[h]
    \small
    \centering
    \resizebox{\linewidth}{!}{
    \begin{tabular}{ccc}
    \toprule
    \textbf{Function} & \textbf{Returns} &\textbf{Description} \\ \midrule
        %(\textbf{AND} u1 u2) & \shortstack{\textbf{u1}: a set of entities \\ \textbf{u2}: a set of entities} & a set of entities & \textbf{AND} function returns the intersection of two arguments \\ \midrule
         (\textbf{AND} u1 u2) & a set of entities & \textbf{AND} function returns the intersection of two arguments \\ \midrule
        (\textbf{COUNT} u) & a singleton set of integer & \textbf{COUNT} function returns the cardinality of the argument \\\midrule
        (\textbf{R} b) & a set of (entity, entity) tuples & \textbf{R} function reverse each binary tuple (x, y) in the input to (y, x) \\\midrule
         (\textbf{JOIN} b u)  
        %  \begin{tabular}[t]{l}\textbf{b}: a set of (entity, entity) tuples \\ \textbf{u}: a set of entities \end{tabular}& 
         & a set of entities  & Inner join based on items in u and the second element of items in b\\ \midrule
         (\textbf{JOIN} b1 b2) 
        %  & \begin{tabular}[t]{l} \textbf{b1}: a set of (entity, entity) tuples \\ \textbf{b2}: a set of (entity, entity) tuples \end{tabular}
        & a set of (entity, entity) tuples & \begin{tabular}[t]{l}Inner join based on the first element of items in b2 and the second \\ element of items in b1\end{tabular}  \\ \midrule
         \begin{tabular}[t]{l}(\textbf{ARGMAX} u b) \\ (\textbf{ARGMIN} u b)\end{tabular}
        %  & \begin{tabular}[t]{l}\textbf{u}: a set of entities \\ \textbf{b}: a set of (entity, value) tuples\end{tabular}
        & a set of entities & Return x in u such that (x, y) is in b and y is the largest / smallest\\\midrule
         \begin{tabular}[t]{l}(\textbf{LT}  b n) \\ (\textbf{LE} b n) \\ (\textbf{GT}  b n) \\ (\textbf{GE} b n)\end{tabular}
        %  & \begin{tabular}[t]{l}\textbf{b}: a set of (entity, value) tuples \\ \textbf{n}: a numerical value\end{tabular}
         & a set of entities & Return all x such that (x, v) in b and v $<$ / $\leq$ / $>$ / $\geq$ n  \\
         \bottomrule
    \end{tabular}}
    \caption{Functions defined in our S-expressions; u: a set of entities, b: a set of (entity, entity) tuples, n: a numerical value.}
    \label{table:lisp}
    \vspace{-10pt}
\end{table}

\nop{
\subsection{Error Analysis}\label{sec:appendix_err}
We also analyze the error rate of our dataset to access the noise level. From 100 randomly sampled logical form templates, their canonical questions and the associated paraphrases, 3 of the 100 canonical questions have mismatching meaning with the logical form, and 12 of the 576 paraphrases don't match the corresponding canonical question, leading to a 3\% error rate of canonical question annotation, 2.1\% error rate of paraphrasing, and 5.6\% error rate overall. All the examined paraphrases are reasonably fluent. 
}

% \begin{figure}[!h]
% \centering
% \includegraphics[width=\linewidth]{figs/distribution.pdf}
% \caption{Distribution of first 3 words in questions. Empty colored blocks are due to either suffixes that are too rare to show individually or lack of space.}
% \label{fig:dist}
% \end{figure}
\textfloatsep 0.15in
\section{Detailed Information of Table~\ref{table:dataset}}
\label{sec:appendix_comparison}
It is not trivial to make statistics of question types, relations, classes, entities and relations for all datasets. Retrieving all those information from \GraphQ and \SimpleQ is straightforward as those information are almost explicitly given in these two datasets. \WebQSP also provides inferential chain that connects the topic entity and the answer entity together with a list of other constraints. Based on these information, we are able to retrieve all the information with no much effort. On the other hand, \QALD and \ComplexQ only provide SPARQL queries with no structured information. It's a little bit challenging to get the information from them. Specifically, we rely on hand-crafted rules (including regular expressions) to retrieve entities, relations and literals. Then we rely on SPARQL algebra to determine whether two question fall into the same type after replacing all entities and literals with placeholders. Also, not all datasets have explict class constraint in their queries like our dataset, so to count the number of different classes, we simply take into account the domain class and range class of all relations in the dataset.
\vspace{-0.1in}
\section{Detailed Experimental Setup}
\label{sec:appendix_setup}
\vspace{-0.05in}

%\subsection{Implementation Details}
%All models in our paper are implemented using AllenNLP. We choose LSTM for both encoder and decoder. The input embeddings and output embeddings are tied during decoding. For BERT-based models, when concatenating the schema items, we choose ';' as delimiter instead of '[SEP]' which is commonly used in text-to-sql since in our experiments we found that using ';' can give us around 3 percent gain of exact match over using '[SEP]'.
% \subsection{Device Information}

% We run all our experiments on a sever with 8 2080ti GPUs. 

% \subsection{Number of Trainable Parameters}

% For BERT-based models, there are 121,291,008 trainable parameters in total. For GloVe-based models, there are 9,394,476 trainable parameters in total as we fix the pre-trained word embeddings during training.
% \subsection{Implementations}
% All models in our paper are implemented using AllenNLP~\cite{gardner-etal-2018-allennlp}. The version of AllenNLP we use is 0.9.0. It provides an implementation for Seq2Seq and also interfaces for both GloVe and BERT. For GloVe-based models, we choose GloVe-6B-300d, and we use BERT-base-uncased as our BERT version. 
\subsection{Hyper-parameters}
We train GloVe-based models with batch size 32 and an initial learning rate of 0.001 using Adam optimizer. For BERT-based models, we are only able to train our model with batch size 1 due to the memory consumption, so we choose a workaround to set the number of gradient accumulation to be 16. We also use Adam optimizer with an initial learning rate of 0.001 to update our own parameters in BERT-based models. For BERT's parameters, we fine-tune them with a learning rate of 2e-5. For all models, the hidden sizes of both encoder and decoder are set to be 768. All hyper-parameters are selected based on the validation set.
\vspace{-0.05in}
\subsection{Training Scheme} 
We use early stopping to train all our models and choose the models with highest exact match on the validation set. The patience is set as 3. In total, we train 2 different models, i.e. GloVe-based Seq2Seq model and BERT-based Seq2Seq model. Note that, one difference in training GloVe-based model is that there is no need to do vocabulary pruning during training.

% \begin{table}[h]
%     \centering
%     \resizebox{0.45\textwidth}{!}{\begin{tabular}{ccc}
%         \toprule
%         &\textbf{In-domain} & \textbf{Cross-domain} \\ \midrule
%         \textbf{Seq2Seq+GloVe+VP} & 0.638 & 0.041 \\
%         \textbf{Seq2Seq+BERT+VP} & 0.658 & 0.327\\
%          \bottomrule
%     \end{tabular}}
%     \caption{Validation Exact Match. }
%     \label{tab:val}
% \end{table}

% \subsection{Transfer Learning on \WebQSP}
% To conduct experiments on \WebQSP, we first convert SPARQL queries in \WebQSP to our S-expression logical forms. To calculate the F1 score of predicted answers, we first convert the predicted logical forms back in to SPARQL queries and then execute the queries against our Virtuoso endpoint to retrieve answers. For pre-training on our dataset, we simply use the parameters trained in the above-mention way to initialize the model, and then we do fine-tuning with the same learning rate and batch size as our main experiments.

\nop{
\subsection{SPARQL Endpoint}
We set up a Virtuoso database to provide a SPARQL endpoint, and the queries for vocabulary pruning and candidates generation and made throught that endpoint. Executing SPARQL queries online is very time-consuming. For experimental purpose, we offline cached some results of vocabulary pruning and candidates generation for reuse.
}

% \begin{table*}[!h]
%     \small
%     \centering
%     \resizebox{\textwidth}{!}{
%     \begin{tabular}{ccccccc}
%     \toprule
%          & s2s+BERT & s2s+BERT+VP & Ranking+BERT & s2s+GloVe & s2s+GloVe+VP & Ranking+GloVe  \\\hline
%          Running time (s)& $4.787\pm0.690$ & $60.899 \pm 47.810$ & $115.459\pm107.866$ & $1.932\pm0.585$ & $50.176 \pm 47.890$ & $80.892 \pm 55.272$\\
%          \bottomrule
%     \end{tabular}}
%     \caption{Running time of different baselines.}
%     \label{table:time}
% \end{table*}

% \section{Efficiency Analysis}
% \label{sec:appendix_efficiency}

% We report the runtime of inference on 1,000 random questions for the models in Table~\ref{table:time}. As we can see, all models with VP or ranking take around 1-2 minutes to process a single query, which is too much of a latency for a production system. This is because both vocabulary pruning and candidate logical forms generation require to run SPARQL queries to get the 2 hop context of topic entities. A promising direction is to develop better search space pruning strategy and integrate that into neural models instead of being separate stages in a pipeline.

\end{document}
\endinput